# Broken-symmetry shape discrimination on a driven Duffing ring

Kaspar A. Schindler

*Sleep-Wake-Epilepsy-Center, Department of Neurology, Inselspital, Bern University Hospital, University of Bern, Bern, Switzerland*
*ORCID: 0000-0002-2387-7767*
*Email: kaspar.schindler@insel.ch*

May 2026

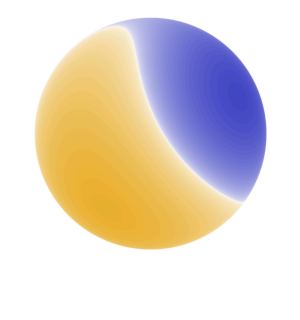

wavecomputing.org

**ABSTRACT**

Distributed computational substrates rely on two elementary operations: *bundling*, the act of populating a shared physical medium with independently retrievable components, and *binding*, the act of composing components into outputs whose identity depends on their relations. We study these two primitives on the simplest closed substrate carrying a continuous symmetry, a cycle graph of $N$ nodes, in two parameter regimes of a single master equation of motion. The linear regime sorts a temporal input across the substrate's $U(1)$-organised eigenmodes, providing a feature representation that matches a windowed-FFT baseline at high signal-to-noise ratio and modestly outperforms it for transient signals at low SNR. The Duffing regime activates a cubic mode-mixing operation constrained by the substrate's symmetry into a sparse selection rule on integer wavenumbers, generating shape-dependent harmonic content that the linear regime cannot produce. We identify a single-number observable, $\phi_0$, that summarises the bound representation's response to input shape, and we analyse its symmetry structure: a $\pi$-periodicity in the shape parameter is exact, while a time-reversal symmetry that would render $\phi_0$ degenerate is broken by the substrate's dissipation. The asymmetric status of these two symmetries is what licenses $\phi_0$ as a meaningful single-number observable; its trajectory across the quotient domain encodes the joint response of binding and dissipation to the input shape. Numerical experiments confirm that $\phi_0$ retains its information content under additive band-limited noise, with seed-averaged means staying clearly above the symmetric-attractor value down to 0 dB input SNR. The framework is developed on synthetic signals only; extensions to richer substrates, more elaborate drives, and real biological signals are open questions for the work that follows.

# 1 Introduction

Reading information out of a physical medium that carries many things at once requires two elementary operations — first a way of separating what is there into independent components, then a way of combining components in structured ways to form outputs that depend on their relations. The two operations have specific names in the literature on distributed representations: *bundling*, the act of populating a shared substrate with several inputs that remain individually retrievable, and *binding*, the act of composing inputs into outputs whose identity depends on relationships among the constituents rather than on their independent presence (Smolensky 1990; Plate 2003; Schindler and Rahimi 2021; Kleyko et al. 2023). Bundling supports separability: a substrate that bundles can store many things and pull each out again. Binding supports compositional structure: a substrate that binds can produce outputs whose informational content lies in the relations among inputs, not just in their individual identities. The two operations together are what allows a physical medium to serve as a substrate for non-trivial information processing rather than as a passive store.

A natural question — and the one that motivates this paper — is to ask what the simplest physical system is in which both operations can be analyzed cleanly. Linear waves on a closed substrate carry an obvious bundling structure: each eigenmode of the substrate is one independent channel, and a temporal input projects onto the eigenmode basis without any combination of channels being performed by the substrate itself. Adding a nonlinearity to such a substrate turns on binding: the nonlinearity mixes input components into outputs at sums and differences of the input frequencies and wavenumbers, with the algebra of the mixing determined by what the substrate's symmetries permit. Both operations are therefore available in a single physical system, and the relationship between them — which the substrate's parameters select between or combine — is what we propose to study.

The cycle graph is the simplest substrate on which this can be done analytically. Its eigenmodes are the discrete Fourier modes of a ring; its continuous symmetry is $U(1)$, the rotations of a circle, which organises the eigenmodes into cosine–sine pairs and constrains the binding algebra into a tractable selection rule on integer wavenumbers modulo $N$. Every other choice of closed substrate carrying a continuous symmetry — and there are many — generates a richer eigenmode structure and a more elaborate binding algebra, but at the cost of having to track more degrees of freedom simultaneously. Starting on the ring isolates the bundling and binding primitives from any further geometric complication, lets the symmetry analysis proceed with minimum apparatus, and gives a clean baseline against which the additional structure of richer substrates can later be measured.

The specific question we use to exercise both primitives is waveform-shape discrimination. Two periodic signals with identical magnitude spectra can have visibly different waveform shapes — one peaked, one sawtoothed, one arched — and these differences are encoded in the relative phases among the harmonics of a fundamental frequency (Cole and Voytek 2017). A measurement tradition in cortical electrophysiology has developed time-domain methods for characterising these waveform features on a cycle-by-cycle basis (Cole and Voytek 2019; Cole 2018; Donoghue, Schaworonkow, and Voytek 2022). A power-spectrum-only analysis is blind to the difference because power is invariant under harmonic phase shifts; a system that preserves only single-frequency response is similarly blind for the same reason. Reading shape requires a substrate that mixes harmonics into outputs whose amplitudes depend on their relative phases, which is exactly the kind of operation that binding through a smooth nonlinearity performs. Whether the bound representation is informative, how it is structured by the substrate's symmetries, and whether it survives realistic noise, are the questions the paper develops.

Three contributions follow. First, we show that the cycle graph in two parameter regimes — linear and Duffing — implements bundling and binding cleanly, with the linear regime sorting inputs across distinguishable channels at temporal resolution set by the channels' own dynamics and the Duffing regime generating shape-dependent harmonic content that the linear regime cannot produce. Second, we identify a single-number observable, $\phi_0$, that summarises the bound representation's response to input shape, and we analyze its symmetry structure: a $\pi$-periodicity in the shape parameter is exact (a property of the equation of motion and the quadratic readout, independent of parameter values), while a time-reversal symmetry that would render $\phi_0$ degenerate is broken by the substrate's dissipation. The asymmetric status of the two symmetries — one exact, one broken — is what makes $\phi_0$ a meaningful single-number observable rather than a fixed constant of the substrate. Third, we show that $\phi_0$ retains its information content under realistic additive noise, with seed-averaged means staying clearly above the symmetric-attractor value down to 0 dB input SNR. What this paper does not do is also worth being explicit about: it does not analyze any real biological signal, it does not propose $\phi_0$ as a clinical biomarker, and it does not compare its performance to specialized shape-detection methods on application data. Those are questions for future work.

The paper is organized as follows. Section 2 specifies the substrate, the master equation of motion in its two parameter regimes, the drive signals, the integration procedure, the eigenmode readout, and the construction of the broken-symmetry observable $\phi_0$. Section 3 develops the bundling, binding, and broken-symmetry framework that interprets the apparatus, with one subsection on each of the three threads plus a brief opening that names them. Section 4 presents the empirical work in three subsections corresponding to the substrate's two regimes plus a noise-robustness analysis. Section 5 situates the contributions in the broader literature on wave-based computation, oscillatory neural dynamics, and reservoir computing, and identifies which features of the analysis are substrate-specific and which generalize. Section 6 concludes; the appendix derives the explicit cubic coupling tensor referenced in Sections 2 and 3.

# 2 Methods

## 2.1 Cycle graph and eigenmodes

The substrate is the cycle graph $C_N$ of $N$ nodes with periodic nearest-neighbour connectivity (Figure 1A). Its graph Laplacian,

$$L = D - A, \tag{1}$$

is the difference of the degree matrix $D$ ($D_{ii} = 2$ for all $i$) and the circulant adjacency matrix $A$ ($A_{ij} = 1$ if and only if $|i - j| \equiv 1 \pmod N$). Because $L$ is circulant, its eigenvectors are the discrete Fourier modes

$$[v_n]_j = \frac{1}{\sqrt{N}} e^{2\pi i n j/N}, \quad n = 0, 1, \ldots, N-1, \tag{2}$$

with eigenvalues

$$\lambda_n = 2(1 - \cos(2\pi n/N)) = 4\sin^2(\pi n/N). \tag{3}$$

We adopt the real-valued basis throughout: each $0 < n < N/2$ supports a degenerate cosine–sine pair $\{c_n, s_n\}$, while $n = 0$ and (for even $N$) $n = N/2$ are non-degenerate. Total eigenmode dimension equals the number of nodes $N$. Representative mode shapes are shown in Figure 1B.

The eigenvalues $\lambda_n = 4\sin^2(\pi n/N)$ are the substrate's intrinsic frequency-squared spectrum: they range from $\lambda_0 = 0$ (uniform mode) to $\lambda_{N/2} = 4$ (Nyquist mode at $n = N/2$). This is the cycle graph's specialisation of the more general graph Fourier transform (Shuman et al. 2013), in which the eigenvectors of a graph Laplacian act as a Fourier basis for signals supported on the graph's nodes and the eigenvalues play the role of graph frequencies. The dynamical mode frequencies $\omega_n$ that the substrate supports under driven dissipative dynamics depend on the parameter regime introduced in §2.2; Figure 1C, which shows the linear-regime dispersion for orientation, is discussed there alongside the equation of motion.

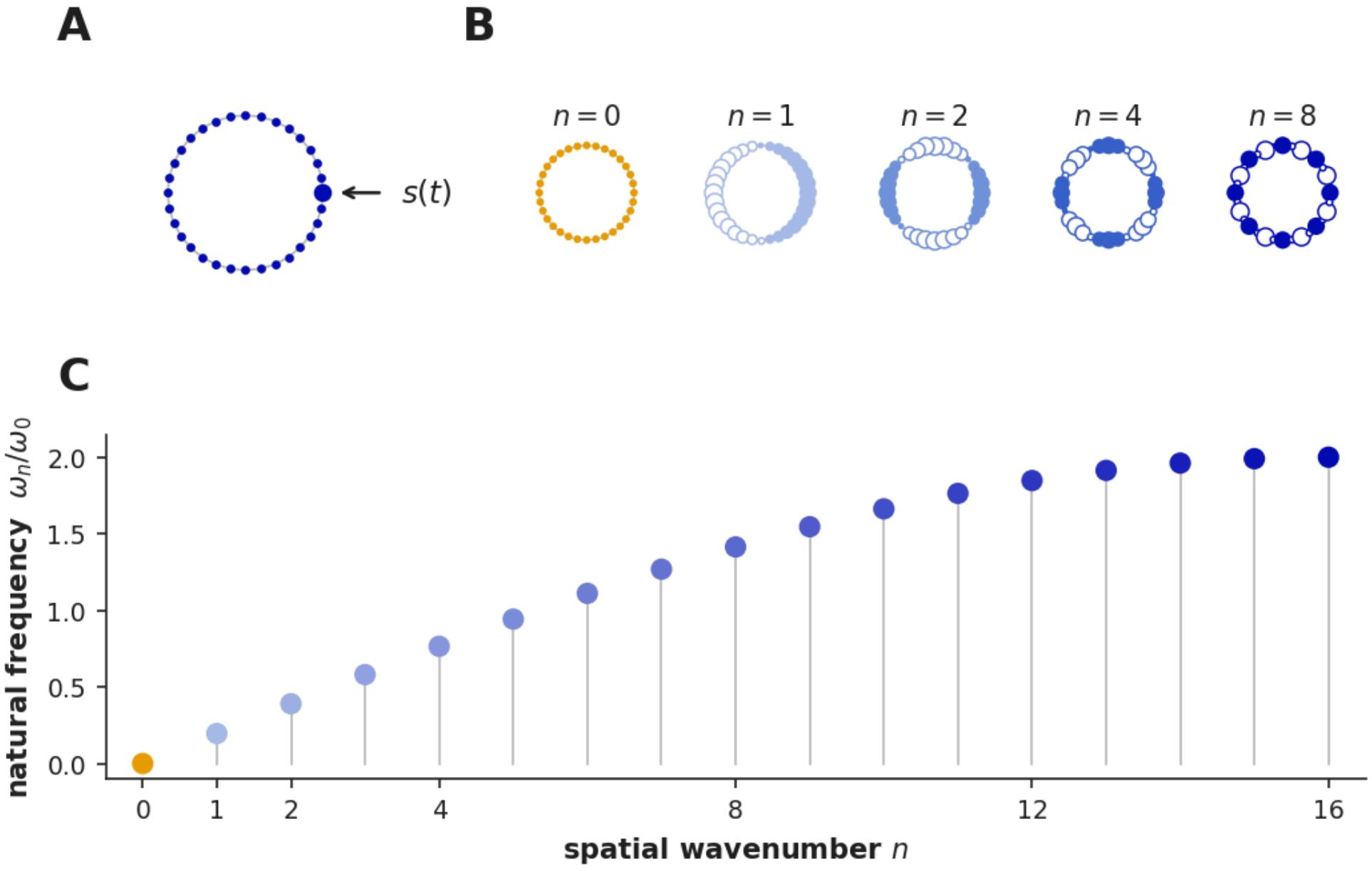

**Figure 1.** Cycle-graph ring substrate ($N = 32$, linear regime). (A) Topology: $N$ nodes in periodic nearest-neighbour connectivity, with drive $s(t)$ injected at node $j = 0$. (B) Representative real-valued Fourier eigenmodes for $n = 0$ (uniform, orange) and $n = 1, 2, 4, 8$ (graded blue); cosine basis shown. (C) Linear-regime dispersion $\omega_n/\omega_0 = \sqrt{\lambda_n} = 2|\sin(\pi n/N)|$, with $\omega_0 \equiv \sqrt{K_c}$ in this regime. The dispersion vanishes for the uniform mode and saturates at $\omega_n/\omega_0 = 2$ at the Nyquist mode $n = N/2$.

## 2.2 Equation of motion: master EOM and two parameter regimes

Each node $i$ of the cycle graph carries a scalar displacement $x_i(t)$ that evolves under the master equation of motion

$$\ddot{x}_i + \gamma \dot{x}_i + \omega_0^2 x_i + \alpha x_i^3 + K_c \sum_j L_{ij} x_j = s_i(t), \tag{4}$$

where $\gamma > 0$ is the damping coefficient, $\omega_0^2$ is the on-site stiffness, $\alpha$ is the Duffing (cubic) nonlinearity, $K_c$ is the nearest-neighbour coupling, and the drive $s_i(t) = s(t)\,\delta_{i,0}$ is injected at the single node $j = 0$; all other nodes are externally undriven.

The four rate parameters $\{\gamma, \omega_0^2, \alpha, K_c\}$ define two regimes used in this paper, distinguished by which terms of Equation 4 are active and by the absolute scales of the parameters that remain.

*Linear regime (Figures 1–3).* We set $\omega_0^2 = 0$ and $\alpha = 0$, with $K_c$ as the sole frequency scale; numerical values $N = 32, \gamma = 0.5$ rad/s, $K_c = (2\pi)^2$ rad²/s². Equation 4 reduces to the bare wave equation on the cycle graph,

$$\ddot{x}_i + \gamma \dot{x}_i + K_c \sum_j L_{ij} x_j = s_i(t). \tag{5}$$

In the eigenmode basis $u_n(t) = \left[V^T x(t)\right]_n$, the dynamics decouple completely:

$$\ddot{u}_n + \gamma \dot{u}_n + \omega_n^2 u_n = s_n(t), \quad \omega_n^2 = K_c \lambda_n, \tag{6}$$

where $s_n(t) = \left[V^T s(t)\right]_n$ is the projection of the drive onto mode $n$. The linear-regime dispersion is therefore $\omega_n = \sqrt{K_c}\sqrt{\lambda_n}$, ranging from 0 (uniform mode) to $2\sqrt{K_c} = 4\pi$ rad/s (Nyquist). This is the dispersion plotted in Figure 1C, with $\omega_0 \equiv \sqrt{K_c}$ as the convenient frequency scale for that regime. Each mode is an independent damped harmonic oscillator at its natural frequency $\omega_n$; integration is performed in eigenmode coordinates by custom RK4.

*Duffing regime (Figures 4–6).* All four terms of Equation 4 are active; numerical values $N = 64$, $\gamma = 0.15$, $\omega_0^2 = 1.0$, $\alpha = 1.5$, $K_c = 0.35$ (dimensionless time). The on-site stiffness $\omega_0^2 x_i$ anchors each node to a stable rest amplitude $x_i = 0$, providing the harmonic restoring force against which the cubic stiffening $\alpha x_i^3$ acts; without it the wave equation has no preferred displacement and the cubic term has no defined working point. The combination of linear restoring force and cubic stiffening at each node is the classical Duffing oscillator (Holmes 1979; Nayfeh and Mook 1979), here distributed across the cycle graph and coupled by the graph Laplacian.

The linear part of Equation 4 (i.e. $\alpha = 0$ held momentarily) gives Duffing-regime mode frequencies

$$\omega_n^2 = \omega_0^2 + K_c \lambda_n, \tag{7}$$

which range from $\omega_0 = 1$ (uniform mode) to $\sqrt{\omega_0^2 + 4K_c} \approx 1.549$ (Nyquist) at the working values. The cubic term, local in node space, becomes a triple convolution in eigenmode space:

$$\ddot{u}_n + \gamma \dot{u}_n + \omega_n^2 u_n + \alpha T_{n\, m_1 m_2 m_3} u_{m_1} u_{m_2} u_{m_3} = s_n(t), \tag{8}$$

with summation implied over $m_1, m_2, m_3$. The cubic coupling tensor $T_{n\, m_1 m_2 m_3}$ inherits the cycle-graph $U(1)$ structure: it is non-zero only for index combinations satisfying the selection rule

$$\pm m_1 \pm m_2 \pm m_3 \equiv n \pmod{N}, \tag{9}$$

with all sign combinations in general allowed. Equation 8 maps temporal harmonics of the drive — generated through the cubic nonlinearity — onto a structured pattern of excited eigenmodes; the explicit form of $T_{n\, m_1 m_2 m_3}$ is derived in Appendix A. Integration is performed in node coordinates by `solve_ivp` (see §2.4 for solver settings per figure).

*Dissipation.* In both regimes $\gamma > 0$ ensures every transient decays at rate $\gamma/2$, so the steady-state response in the analysis window of §2.5 is well-defined and free of initial-condition memory.

## 2.3 Drive signals

The drive in Equation 4 is injected at the single node $j = 0$ as $s_i(t) = s(t)\,\delta_{i,0}$. The waveform $s(t)$ is regime-specific.

*Linear regime: substrate-characterisation battery (Figure 2).* Four canonical drives exercise the substrate's spectral and temporal response: a pure tone $s(t) = \cos(\omega_5\, t)$ at the natural frequency of mode $n = 5$; a linear chirp $s(t) = \cos\left[\omega_1 t + \frac{1}{2}(\omega_{12} - \omega_1)\, t^2/T\right]$ sweeping the instantaneous frequency from $\omega_1$ to $\omega_{12}$ over the simulation duration $T = 16$ s; a Gaussian burst $s(t) = \exp\left[-(t - t_c)^2/(2\sigma^2)\right]\cos(\omega_8\, t)$ at $t_c = 8$ s with envelope width $\sigma = 1.2$ s, carrying energy at mode $n = 8$ but localised in time; and a frequency-modulated tone $s(t) = \cos[\omega_8\, t + (D/\omega_m)\sin(\omega_m\, t)]$ with carrier $\omega_8$, modulation rate $\omega_m = 0.4$ rad/s, and modulation depth $D = 0.25$, exercising narrow-band temporal structure embedded in a single eigenmode resonance.

*Linear regime: classification battery (Figure 3).* A four-class problem distinguishes a noise-only class from three pure-tone classes at well-separated wavenumbers $n \in \{3, 7, 11\}$. Each trial is constructed as $x(t) = \cos(\omega_n\, t + \varphi_0) + A_{\text{noise}}\, \eta(t)$ for the tone classes (with $\varphi_0$ uniform on $[0, 2\pi)$ per trial and $\eta(t)$ unit-variance white Gaussian noise), or just $A_{\text{noise}}\, \eta(t)$ for the noise class. The noise amplitude $A_{\text{noise}} = 10^{-\text{SNR}/20}$ sets the input SNR with signal amplitude fixed at unity (an amplitude-SNR convention; the Duffing-regime noise extension below uses a power-SNR convention, the two differing by a factor of two in dB); eleven SNR values in $[-24, 0]$ dB are sampled.

*Duffing regime: two-tone shape probe (Figures 4–5).* The drive is the two-tone signal

$$s(t) = A_1 \cos(2\pi f_{\text{drive}}\, t) + A_2 \cos(4\pi f_{\text{drive}}\, t + \Delta\phi_2), \tag{10}$$

with $A_1 = 0.6, A_2 = 0.30, f_{\text{drive}} = 0.18$ (dimensionless), and $\Delta\phi_2$ the relative phase of the second harmonic. At fixed amplitudes, $\Delta\phi_2$ alone controls the waveform shape: $\Delta\phi_2 = 0$ yields an amplitude-asymmetric peaked shape, $\Delta\phi_2 = \pi/2$ a time-asymmetric sawtooth, and the full sweep $\Delta\phi_2 \in [0, 2\pi)$ traces the shape manifold. Crucially, the magnitude spectrum $|c_k|^2$ of $s(t)$ is independent of $\Delta\phi_2$, so any $\Delta\phi_2$ -dependence in the reservoir's response is a genuine shape signature rather than a residual power-spectrum effect.

*Duffing regime: noise extension (Figure 6).* For the noise-robustness test, additive Gaussian noise is added to Equation 10: $s(t) \rightarrow s(t) + n(t)$. The noise process is constructed by low-pass-filtering white Gaussian samples through a 4th-order Butterworth filter with cutoff $f_c = 5$ (dimensionless, $\sim 5\, f_{\text{drive}}$), then rescaling to a target standard deviation $\sigma_{\text{noise}} = \sigma_s/\sqrt{10^{\text{SNR}/10}}$, where $\sigma_s = \sqrt{(A_1^2 + A_2^2)/2}$ is the RMS amplitude of the clean two-tone (a power-SNR convention). Four SNRs — 30, 20, 10, and 0 dB — are reported, each across 8 independent random seeds.

## 2.4 Numerical integration and noise

The two regimes use different integration strategies, each chosen to exploit the regime's structure.

*Linear regime.* The eigenmode EOM (Equation 6) decouples by mode, so each mode's two-dimensional phase-space ODE is integrated independently using fixed-step RK4. Timestep $\Delta t = 0.01$ s, simulation duration $T = 16$ s ($N_T = 1600$ samples), initial conditions $u_n(0) = \dot{u}_n(0) = 0$ for all $n$. RK4 is preferred over adaptive methods here: the linear, decoupled modes are well conditioned, and the explicit form avoids any solver-state coupling between modes.

*Duffing regime.* The on-site cubic and graph-Laplacian terms couple all $N$ nodes, so Equation 4 is integrated in node coordinates using `solve_ivp` (`scipy.integrate`). Initial conditions are $x_i(0) = \dot{x}_i(0) = 0$ for all $i$; output is sampled at $\Delta t = 0.05$ (dimensionless time). Solver method, tolerances, and total simulation time vary by figure (Table 1). Tolerances are tightest for F5 (DOP853, rtol $= 10^{-10}$, atol $= 10^{-12}$) where the symmetry analysis demands recovery of $\phi_0$ to relative precision $\sim 10^{-4}$, and loosest for F6 (RK45, rtol $= 10^{-6}$, atol $= 10^{-8}$) where the noise dominates the per-step error budget and tighter tolerances would only track fluctuations the readout integrates out.

*Steady-state window (Duffing regime).* For each Duffing-regime simulation, the first half of the trajectory is discarded as transient and the analysis window is the largest integer number of drive periods that fits in the second half:

$$n_{ss} = \lfloor (T_{tot}/2)\,/\,T_{period} \rfloor, \quad \mathrm{SS_LEN} = n_{ss} \cdot T_{period}/\Delta t, \tag{11}$$

where $T_{period} = 1/f_{drive}$. This places every drive harmonic on a single FFT bin — eliminating spectral leakage in the harmonic-energy readout (§2.5) — and gives every reported $E_k$ a consistent denominator. For the figures used in this paper, $T_{tot}$ is chosen so that $(T_{tot}/2)/T_{period}$ is exactly integer (Table 1).

*Noise integration (Figure 6).* The band-limited noise process $n(t)$ defined in §2.3 lives on the storage grid $\{k\Delta t\}_{k=0}^{T_{tot}/\Delta t}$. The adaptive solver evaluates the drive at arbitrary intermediate times, so $n(t)$ is queried via cubic-spline interpolation between grid points, preserving smoothness ($C^2$) without aliasing artefacts at the Butterworth cutoff $f_c = 5$. Each (SNR, seed) pair generates an independent realisation of $n(t)$ with the procedure of §2.3; 32 shape-phase points $\Delta\phi_2 \in [0, 2\pi)$ are then swept per realisation, yielding $4 \times 8 \times 32 = 1024$ Duffing-regime simulations underlying Figure 6.

## 2.5 Eigenmode readout and harmonic energies

The reservoir output is the projection of the node trajectory onto the eigenmode basis,

$$a_n(t) = \sum_j V_{jn}\, x_j(t), \tag{12}$$

giving one real-valued time series per eigenmode. In the linear regime, the integration coordinates are exactly the mode amplitudes (§2.2); in the Duffing regime, the projection (Equation 12) is applied at each output sample of the node-coordinate solver. The two regimes use different aggregations of $\{a_n(t)\}$, matched to the figure being produced.

*Linear regime: Hilbert-envelope features (Figures 2, 3).* For each mode we form the analytic-signal envelope $|a_n(t)|_{env} = |a_n(t) + i\,\mathcal{H}\{a_n\}(t)|$ via the Hilbert transform $\mathcal{H}$, suppressing the carrier oscillation at $\omega_n$ and exposing the slow modulation of each

| Fig. | Regime | $N$ | Method | rtol / atol | $T_{tot}$ | $n_{ss}$ |
|---|---|---|---|---|---|---|
| 2, 3 | Linear | 32 | RK4 (fixed) | — | 16 s | — |
| 4 | Duffing | 64 | RK45 | $10^{-8}$ / $10^{-10}$ | 400 | 36 |
| 5 | Duffing | 64 | DOP853 | $10^{-10}$ / $10^{-12}$ | 300 | 27 |
| 6 | Duffing | 64 | RK45 | $10^{-6}$ / $10^{-8}$ | 200 | 18 |

**Table 1.** Numerical integration settings per figure. Linear-regime figures use fixed-step RK4 in eigenmode coordinates; Duffing-regime figures use adaptive `solve_ivp` in node coordinates. $n_{ss}$ is the number of drive periods retained as the steady-state analysis window (Equation 11), defined only for the Duffing regime.

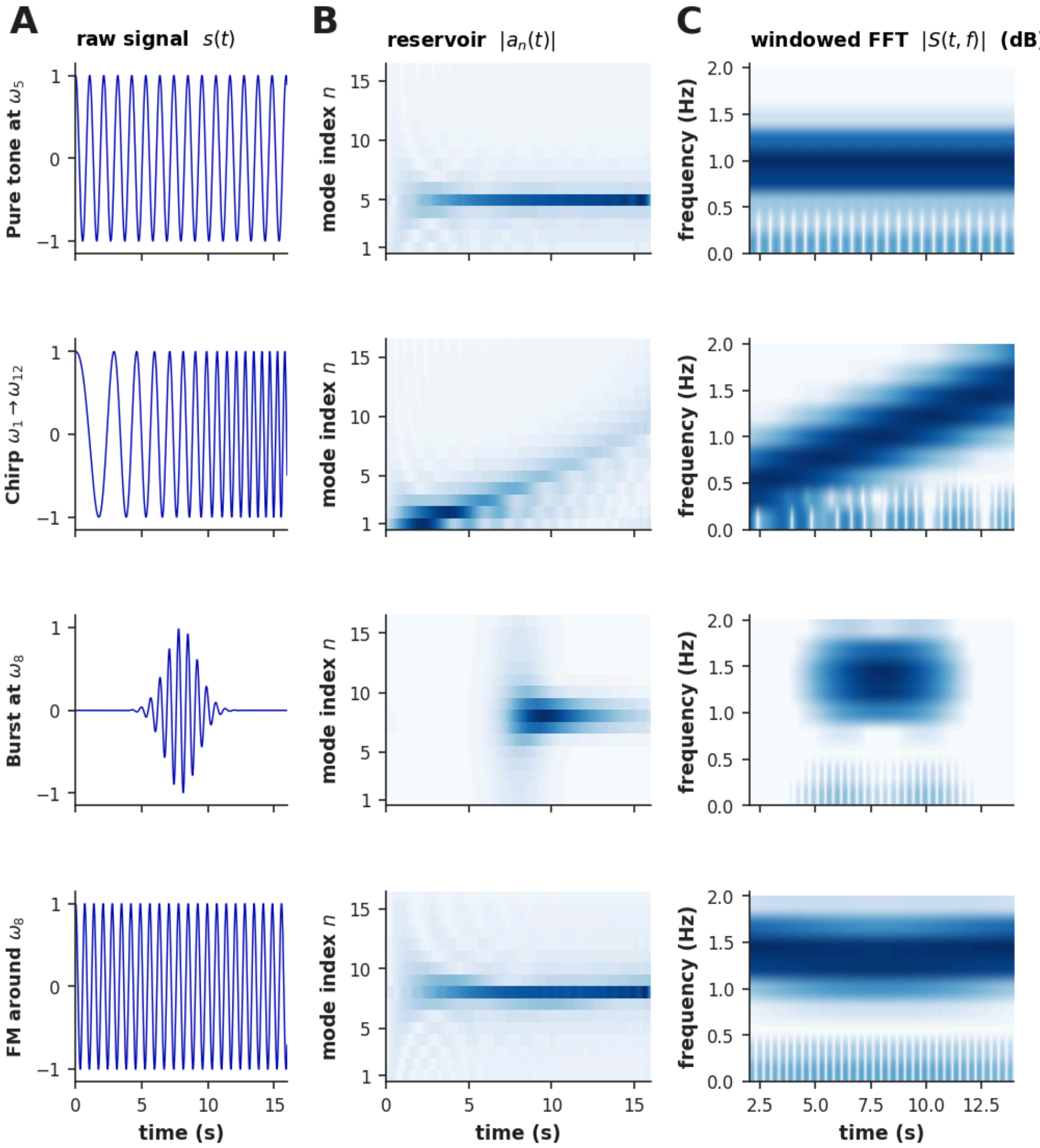

**Figure 2.** Eigenmode reservoir matches windowed FFT on stationary signals and exceeds it on transients (linear regime, $N = 32, \gamma = 0.5$ rad/s). Each row shows a canonical drive signal (left), the reservoir's real-valued mode amplitude $|a_n(t)|$ (centre), and the windowed-FFT spectrogram $|S(t, f)|$ in dB (right). Top to bottom: pure tone at $\omega_5$; chirp from $\omega_1$ to $\omega_{12}$; Gaussian burst at $\omega_8$; FM around $\omega_8$. The reservoir tracks the burst's mode-localised signature with higher temporal resolution than the FFT spectrogram while matching it on the stationary tone.

mode's amplitude. For Figure 2, $|a_n(t)|_{\text{env}}$ is plotted as a (mode index, time) heatmap. For the classifier in Figure 3, the feature vector for each trial is the mode-wise mean of the envelope over the second half of the trial,

$$f_n^{\text{res}} = \langle |a_n(t)|_{\text{env}} \rangle_{t>T/2}, \tag{13}$$

discarding the pre-stationary transient and giving a 16-dimensional descriptor (one entry per non-trivial mode for $N = 32$).

*Linear regime: windowed-FFT baseline.* For comparison we apply a windowed FFT directly to the input signal $x(t)$, bypassing the reservoir entirely. `scipy.signal.spectrogram` is used with a Hann window of 4 s default duration (1 s for the chirp display panel only) and 50 ms hop in Figure 2, and with 75 % overlap and time-averaged across all bins in Figure 3. For the classifier baseline the time-averaged spectrum is then sampled at the eigenmode frequencies $\omega_n / 2\pi$, giving a matching 16-dimensional FFT feature vector $f^{\text{fft}}$ per trial.

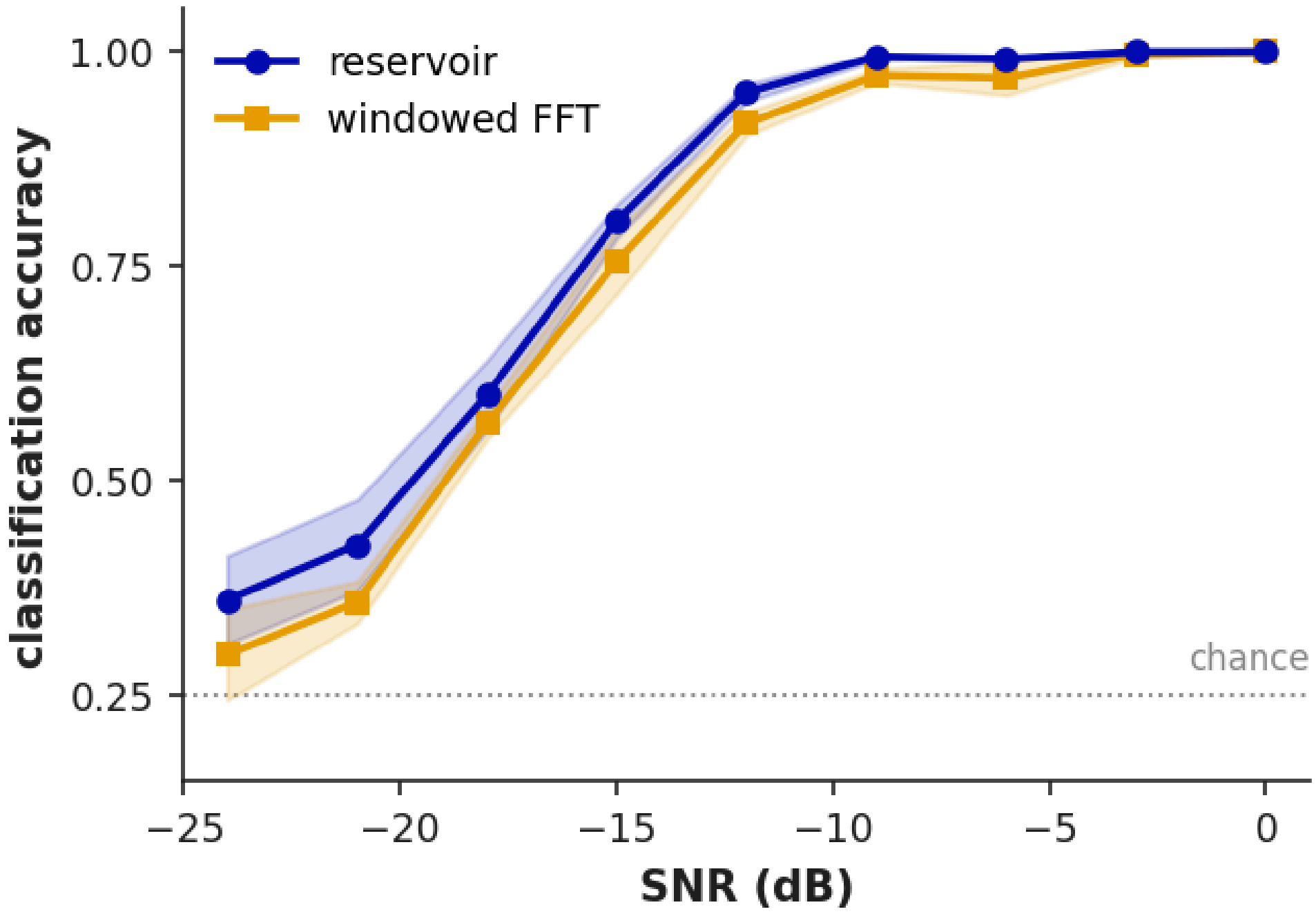

**Figure 3.** Weak-signal classification: reservoir vs windowed-FFT baseline (linear regime). Classification accuracy across the four canonical drive classes plus a noise class as a function of input SNR. Reservoir features (blue circles) maintain $> 90\%$ accuracy down to SNR $\approx -12$ dB; the windowed-FFT baseline (orange squares) crosses the same threshold $\sim 3$ dB earlier. Shaded bands are $\pm 1$ standard deviation across cross-validation folds; dashed grey line marks chance $(25\%)$.

*Linear regime: ridge classifier (Figure 3).* Both feature vectors $f^{\text{res}}$ and $f^{\text{fft}}$ are fed to the same linear classifier trained by ridge regression with regularisation $\lambda = 10^{-3}$ against one-hot class targets; predictions are taken as the argmax of the output score vector. Each SNR is evaluated on 200 training and 200 held-out test trials drawn independently per condition.

*Duffing regime: harmonic energies $E_k$ (Figures 4–6).* The shape-discrimination readout is the energy in the $k$-th temporal harmonic of the drive, summed over the spatial response,

$$E_k = \sum_j \left| \hat{x}_j(k\,\omega_{\text{drive}}) \right|^2, \tag{14}$$

where $\omega_{\text{drive}} = 2\pi f_{\text{drive}}$ and $\hat{x}_j(\omega)$ is the discrete Fourier transform of $x_j(t)$ over the steady-state window of §2.4. By Parseval's theorem and orthogonality of $\boldsymbol{V}$, the node-summed and mode-summed forms agree: $\sum_j \left|\hat{x}_j\right|^2 = \sum_n |\hat{a}_n|^2$, so $E_k$ is a spatial invariant of the ring's response and independent of which node carries the drive. The steady-state window choice (Equation 11) places each harmonic on a single FFT bin at integer index $k\, n_{\text{ss}}$, so Equation 14 is evaluated by direct bin look-up with no leakage correction. We report harmonics $k = 1, \ldots, 6$. The shape observable $\phi_0$ derived from $E_5(\Delta\phi_2)$ is defined in §2.6.

### 2.6 The shape observable $\phi_0$

The harmonic energies $E_k(\Delta\phi_2)$ from §2.5 characterise the Duffing ring's response to the two-tone shape probe of §2.3. Among $k = 1, \ldots, 6$, the fifth harmonic shows the largest shape contrast at the working point $\alpha = 1.5$ (Figure 4C), and we therefore use $E_5(\Delta\phi_2)$ as the shape-coordinate function. The shape observable is defined as the location of its peak,

$$\phi_0 \equiv \arg \max_{\Delta\phi_2 \in [0,\pi)} E_5(\Delta\phi_2). \tag{15}$$

*Domain* $[0, \pi)$. The natural sweep is over $\Delta\phi_2 \in [0, 2\pi)$, but $E_5$ satisfies the exact symmetry $E_5(\Delta\phi_2 + \pi) = E_5(\Delta\phi_2)$ (*Symmetry II* derived in §3.4). The two peaks at $\phi_0$ and $\phi_0 + \pi$ are therefore identified, and the natural domain of the shape observable is the quotient $[0, \pi)$. By contrast, the time-reversal *Symmetry I*, $E_5(-\Delta\phi_2) = E_5(\Delta\phi_2)$, is broken by the damping term in Equation 4; the peak position therefore drifts away from the conservative-attractor values $\{0, \pi/2, \pi\}$ as the damping-driven phase lag accumulates with increasing $\alpha$, sweeping a continuous trajectory across the quotient domain (Figure 5C).

*Sampling and trigonometric interpolation.* $E_5$ is sampled on a uniform grid of $N_\phi$ shape-phase points $\Delta\phi_2^{(m)} = 2\pi m/N_\phi$, $m = 0, \ldots, N_\phi - 1$, with $N_\phi = 64$ for the noise-free sweep (Figure 5A) and $N_\phi = 32$ per noise realisation in Figure 6. The $N_\phi$ samples uniquely determine a $2\pi$-periodic band-limited trigonometric polynomial of order $N_\phi/2$ that interpolates them. We compute its discrete Fourier coefficients $\hat{E}_n = (1/N_\phi) \sum_m E_5^{(m)} \exp(-2\pi i\, nm/N_\phi)$ via `numpy.fft.rfft`, evaluate the reconstruction

$$\tilde{E}_5(\Delta\phi_2) = \hat{E}_0 + 2 \sum_{n=1}^{N_\phi/2-1} \mathrm{Re}\big[\hat{E}_n\, e^{in\,\Delta\phi_2}\big] + \hat{E}_{N_\phi/2} \cos\!\left(\frac{N_\phi}{2}\, \Delta\phi_2\right) \tag{16}$$

on a fine grid of 2048 points in $[0, 2\pi)$, take the argmax, and fold the result into $[0, \pi)$ via Symmetry II. The fine-grid resolution $\Delta\phi_2 \approx 0.003$ ($\sim 10^{-3}\pi$) is well below the seed-to-seed scatter at the SNRs of interest (Figure 6), so the reconstruction step does not limit precision.

*Aggregation under noise.* For each SNR level reported in Figure 6, $\phi_0$ is estimated independently for each of 8 random noise seeds, and we report the seed-wise sample mean and sample standard deviation. Individual seeds are also displayed as faint scatter markers alongside the aggregate, providing visual access to the underlying distribution rather than its summary alone.

## 3 Bundling, binding, and broken symmetry

### 3.1 What this section does

Methods built an apparatus. We have a substrate (the cycle graph), an equation of motion (the master EOM with two parameter regimes), drives (the canonical signals and the two-tone shape probe), a readout (mode amplitudes and harmonic energies $E_k$), and a single observable derived from the readout ($\phi_0$). What this section does is explain why the apparatus works — why the eigenmode decomposition is the right language for the substrate, why the cubic nonlinearity has any business discriminating waveform shape, and why $\phi_0$ in particular carries information rather than being a degenerate or noise-driven quantity.

Two conceptual threads run through the explanation.

The first is computational. The linear ring and the Duffing ring implement two distinct primitives that have specific names and a substantial cross-disciplinary literature: linear superposition of independent eigenmodes is a *bundling* operation, in which multiple inputs occupy a common substrate while remaining individually retrievable; cubic mode-mixing through the Duffing nonlinearity is a *binding* operation, in which input components are composed algebraically into outputs that depend on the relations between them rather than on their independent presence (Schindler and Rahimi 2021). Bundling supports separability; binding supports compositional structure. Reading the paper's two regimes through this lens makes their relationship plain: the linear reservoir bundles, the Duffing reservoir binds, and the harmonic energies $E_k$ are the bound representation of the input's waveform shape.

The second thread is structural. The substrate's symmetries determine which binding algebras are permitted on it, which observables are degenerate, and what changes under physical perturbations like damping. The cycle graph carries the simplest closed continuous symmetry available to a one-dimensional wave substrate — a circle of rotations, $U(1)$ — and this single fact propagates through everything that follows: the two-fold cosine/sine degeneracy of every non-trivial eigenmode, the cubic-mixing selection rule $\pm m_1 \pm m_2 \pm m_3 \equiv n \pmod N$ that we wrote down in Methods (Equation 9) without yet explaining where it comes from, the $\pi$-periodicity that defines the natural domain of $\phi_0$, and the time-reversal symmetry whose breaking by dissipation is what makes $\phi_0$ a non-trivial observable in the first place.

The two threads are intertwined. Symmetry constrains which bindings the substrate can perform; the bindings that get realised are what give the bundled mode amplitudes their compositional content. Subsections 3.2–3.4 develop the threads in turn — bundling on the linear ring, binding through the cubic cascade, and the two waveform symmetries that determine when $\phi_0$ is informative — with both threads visible throughout.

### 3.2 Waves on a ring as bundling substrate

To put information into a substrate, the substrate must offer distinguishable states. This is the elementary requirement that any computational medium satisfies, and the one that organises everything that follows. The cycle graph satisfies it through the eigenmodes of its graph Laplacian: each eigenmode is a spatial pattern that no other eigenmode replicates, and the substrate's representational capacity is set by the number of such patterns it supports.

The eigenmodes are the spatial Fourier basis we wrote in Methods (Equation 2), and they have the structure that one expects of a ring. The $n = 0$ mode is uniform across all nodes — the same value everywhere — and is invariant under every rotation that maps the ring to itself. A uniform pattern is also a maximally symmetric pattern, and a maximally symmetric pattern carries no spatial information: every location on the ring is identical to every other, and the substrate in this state has nothing to say. Each excitation of an $n > 0$ mode breaks this rotational ($U(1)$) symmetry by picking out a particular phase angle around the ring. Symmetry breaking is, in this concrete sense, information creation: the substrate has gone from one indistinguishable configuration to a specific distinguishable one.

The ring's two-fold degeneracy structure follows from the same symmetry. For each non-trivial wavenumber $0 < n < N/2$, the cycle graph supports a cosine mode $c_n \propto \cos(2\pi nj/N)$ and a sine mode $s_n \propto \sin(2\pi nj/N)$ with identical eigenfrequency $\omega_n$; the modes $n = 0$ and (for even $N$) $n = N/2$ are non-degenerate. The two members of each degenerate pair are related by a rotation of the ring through a quarter wavelength, so the pair encodes the same spatial wavelength at two orthogonal phase angles. Total eigenmode dimension equals the number of nodes $N$ — $1 + 2 \cdot (N/2 - 1) + 1 = N$ for even $N$ — and this is the substrate's bundling capacity, the number of independent channels into which a temporal input can be loaded without interference between channels. This is the simplest non-trivial degeneracy that a closed continuous symmetry permits: a rotation about a circle, parameterised by one angle, organises modes into pairs. Symmetry groups with more parameters generate richer degeneracy multiplets, but already on the ring the structure is enough to do useful work.

The bundling that the linear regime performs is the operation of populating these channels. A drive $s(t)$ injected at node $j = 0$ projects onto every eigenmode through the weight $[v_n]_0$, and each eigenmode then evolves as an independent damped harmonic oscillator at its own natural frequency $\omega_n$ (Equation 6). Different temporal frequencies in the drive resonate with different eigenmodes, so a temporally complex input gets decomposed across the substrate into a set of components that occupy the same physical ring without interfering with one another. Mode 5 carries the drive's energy near $\omega_5$, mode 12 carries the drive's energy near $\omega_{12}$, and the two coexist on the ring as cleanly as if they were on separate substrates. The Hilbert-envelope readout of Methods §2.5 is the natural retrieval operation for this kind of bundling: it extracts each channel's amplitude trajectory $|a_n(t)|_{\text{env}}$ without requiring the reservoir to disentangle anything that was not already separable. Figures 2 and 3 show this retrieval working — and working slightly better than a direct windowed-FFT baseline applied to the raw input, because the reservoir's mode-resolved temporal envelope tracks burst-like transients that the FFT's fixed-window analysis smears out.

The diagnostic of bundling is the count of distinguishable states, and on the ring this count is bounded above by $N$. Bundling tells us what is in the substrate without telling us how the components combine. For combination — for the substrate to do something with the relations between its channels rather than treating each independently — a different operation is needed, and the linear ring cannot supply it. That operation is the subject of the next subsection.

### 3.3 Cubic nonlinearity as a binding operation

A linear system has a striking and restrictive property: it maps sines to sines. A pure sinusoidal drive at frequency $\omega$ produces, in any linear damped oscillator, a steady-state response that is itself a sine at frequency $\omega$ — shifted in phase, scaled in amplitude, but containing no other frequency component. The same is true on the ring as a whole: an input drive made of one frequency excites mode amplitudes that oscillate at that one frequency, and the substrate's response carries no temporal information that was not already in the drive. Bundling, in the linear regime, is therefore conservative: the substrate sorts the input across channels but does not generate anything new.

This means that any deviation of a system's steady-state response from a pure sine wave at the drive frequency is a fingerprint of nonlinearity. The fingerprint takes a specific form: harmonics, that is, energy at integer multiples $k\omega$ of the drive frequency. A signal that visibly differs from a sinusoid — a peaked wave, a sawtooth, an arch — is a signal whose spectrum contains energy at $2\omega$, $3\omega$, and so on, and these higher harmonics can only have come from a system that did not preserve the single-frequency structure of the drive. What harmonic generation means computationally is that the system has performed an operation on the drive that mixes its components into new ones, producing outputs whose frequencies are arithmetic combinations of the input's frequencies. This kind of operation has a name in the literature on distributed representations: it is a *binding* operation, an algebraic composition that combines several inputs into a single output whose identity depends on the relations among the inputs rather than on their independent presence (Schindler and Rahimi 2021; Cole and Voytek 2017). A binding operation produces a representation that is not just "what is in the substrate" but "how the parts of the input combine."

The Duffing nonlinearity in the master EOM (Equation 4) is the simplest local binding operation a smooth oscillator can perform. The cubic term $\alpha x_i^3$ takes the displacement at one node and generates a new local force whose Fourier expansion contains $3\omega$ alongside $\omega$ when $x_i$ oscillates at $\omega$, and generates $\omega_a + \omega_b$, $\omega_a - \omega_b$, $2\omega_a + \omega_b$, and analogous arithmetic combinations when $x_i$ contains two frequencies. At every node, the cubic produces sums and differences of the frequencies present. This is the entire content of harmonic mixing. The same arithmetic — triples of frequencies and wavenumbers combining into outputs at sums and differences under a cubic nonlinearity — appears across nonlinear physics in three- and four-wave mixing in nonlinear optics and in resonant triad interactions in wave turbulence (Boyd 2020; Zakharov, L'vov, and Falkovich 1992), where the spatial wavenumbers participate in selection rules fixed by phase matching; what the closed substrate adds is to convert these selection rules from continuous phase-matching conditions into the discrete modular arithmetic of a finite eigenmode basis.

What makes the ring's binding operation interesting — what distinguishes it from the same cubic acting in free space — is that the substrate is closed. A wave excited at node $j = 0$ does not propagate off to infinity. It travels around the ring, returns to its origin, and superposes with whatever the drive has injected in the meantime; the wavefront meets itself. This self-encounter is what allows the cubic mixing at one node to produce coherent contributions to a global eigenmode rather than a local broadband mess. A drive component that excites spatial mode $m_1$ and a drive component that excites spatial mode $m_2$ wrap around the ring with their respective spatial periodicities, and where they meet — at every node, in steady state — their cubed superposition produces a contribution at spatial wavenumber $\pm m_1 \pm m_2 \pm m_3$ for any third mode $m_3$ also present, evaluated modulo the ring's circumference. Reading off the algebra:

$$\pm m_1 \pm m_2 \pm m_3 \equiv n \pmod N, \tag{17}$$

which is exactly the selection rule we wrote down in Methods (Equation 9) without yet explaining where it comes from. The mod-$N$ arithmetic is not a formal curiosity — it is the mathematical residue of waves wrapping around a finite closed substrate and meeting themselves. A closed surface is the geometric prerequisite for self-coherent mode mixing; an open chain, however long, would smear the same cubic interaction into a broadband response with no selection-rule structure at all.

The binding operation that the closed ring permits, then, is constrained by its $U(1)$ symmetry into a specific algebra: triples of input wavenumbers compose into output wavenumbers under modular addition with sign freedom, and the cubic coupling tensor $T_{n\,m_1 m_2 m_3}$ that appears in the eigenmode-space EOM (Equation 8) is the substrate-determined multiplication table for this algebra. The harmonic energies $E_k$ that we read out of the Duffing ring (Methods, Equation 14) are therefore not just power measurements; they are the bound representation of the input waveform's shape, distributed across spatial modes by the substrate and across temporal harmonics by the cubic, with the structure of the redistribution dictated by the ring's symmetry. $\phi_0$, defined in Methods as the peak of $E_5(\Delta\phi_2)$, is one specific projection of this bound representation onto a single number — the projection that is most sensitive to the broken-symmetry shift discussed in the next subsection.

A note on what generalises and what is specific. The ring is one example of a closed substrate carrying a continuous symmetry, and the modular-addition selection rule is what that particular symmetry permits. A closed two-dimensional substrate carrying a different continuous symmetry would generate a different algebra of admissible bindings, typically with richer multiplet structure and more elaborate composition rules. The bundling/binding distinction itself — linear superposition populates channels, nonlinearity composes them — is general; the specific algebra depends on the substrate.

### 3.4 Two waveform symmetries and why $\phi_0$ is informative

A short word on what symmetry means in this paper, before we put it to work. The general idea is simple: a symmetry of some object is a transformation that you can apply to that object without changing it. A circle has rotational symmetry because rotating it leaves it looking the same; a square does not have full rotational symmetry, because most rotations turn it into a tilted square that you can tell apart from the original. The pattern is general: if a transformation leaves an object invariant, the states related by that transformation are indistinguishable along whatever axis the transformation acts. This is the second time indistinguishability has appeared in the paper; the first was in Subsection 3.2, where it set the lower bound on what a substrate can encode. Symmetry and information are two faces of the same coin: a symmetry that holds is what makes a set of states indistinguishable in some respect, and breaking that symmetry is what makes them distinguishable again. This is why, throughout this paper, broken symmetries do informational work that intact symmetries cannot.

Two kinds of symmetry need to be kept distinct. The substrate carries its own symmetries — the ring's $U(1)$ rotational symmetry that we have leaned on twice already, in Subsection 3.2 for the eigenmode degeneracy and in Subsection 3.3 for the cubic-mixing selection rule. But the response of a system, viewed as a function of some external parameter we vary, can also be symmetric or asymmetric in that parameter. This second kind of symmetry depends on the dynamics, on the drive, and on which observable we read off, in addition to the substrate. The two symmetries we analyse below are of this second kind: properties of the harmonic energy $E_5(\Delta\phi_2)$ as a function of the second-harmonic phase $\Delta\phi_2$, with everything else — the substrate, the drive amplitudes, the nonlinearity — held fixed. One of these symmetries holds exactly; the other is broken by the system's dissipation. The asymmetric status of the two is what makes $\phi_0$ a non-trivial observable rather than a degenerate one.

*Symmetry II: $\pi$-periodicity in $\Delta\phi_2$.* Consider what happens when we shift the second-harmonic phase by $\pi$. The drive $s(t) = A_1 \cos(\omega t) + A_2 \cos(2\omega t + \Delta\phi_2)$ becomes $s_\pi(t) = A_1 \cos(\omega t) + A_2 \cos(2\omega t + \Delta\phi_2 + \pi) = A_1 \cos(\omega t) - A_2 \cos(2\omega t + \Delta\phi_2)$, that is, the second-harmonic component flips sign while the fundamental is untouched. A $\pi$ shift in the second harmonic and a sign flip of the second harmonic are the same operation, by the cosine identity $\cos(\theta + \pi) = -\cos\theta$. Now consider the same phase shift combined with a half-period time translation $t \to t + T/2$ where $T = 2\pi/\omega$. Under the time translation, $\cos(\omega t)$ picks up a sign flip ($\cos(\omega(t + T/2)) = \cos(\omega t + \pi) = -\cos(\omega t)$), while $\cos(2\omega t + \cdot)$ is unchanged ($\cos(2\omega(t + T/2)) = \cos(2\omega t + 2\pi) = \cos(2\omega t)$). So the combined transformation $(t, \Delta\phi_2) \to (t + T/2, \Delta\phi_2 + \pi)$ flips the sign of the fundamental *and* flips the sign of the second harmonic — in other words, multiplies the entire drive by $-1$.

A drive multiplied by $-1$ is, on the symmetric ring with the master EOM (Equation 4), simply $-x_i(t)$ as the solution: the EOM is invariant under $x_i \to -x_i$ at every node (the linear, damping, and Laplacian terms are all linear; the cubic term sends $x_i^3 \to -x_i^3$, matching the sign flip on the right-hand side). The harmonic energies $E_k = \sum_j \left|\hat{x}_j(k\omega)\right|^2$ are quadratic in $x_i$, so they are insensitive to the sign: $E_k(-x_i) = E_k(x_i)$. Putting these together,

$$E_k(\Delta\phi_2 + \pi) = E_k(\Delta\phi_2) \quad \text{for every } k\text{, exactly,} \tag{18}$$

which is what we have been calling Symmetry II. The symmetry is exact because every step of the argument relies only on the structure of the EOM and the readout, not on any parameter values. Its consequence is concrete: as $\Delta\phi_2$ runs through the sweep $[0, 2\pi)$, the function $E_k(\Delta\phi_2)$ traces every value it will ever take twice, once in $[0, \pi)$ and once again identically in $[\pi, 2\pi)$. The natural domain on which a peak position can carry non-redundant information is therefore the quotient $[0, \pi)$, and any observable that confuses these two halves — by reporting both peaks separately, or by sweeping a region that crosses $\Delta\phi_2 = \pi$ without folding — is reading indistinguishable states as if they were distinguishable.

*Symmetry I: time-reversal in $\Delta\phi_2$.* The second candidate symmetry is reflection of $\Delta\phi_2$ about zero, $E_k(-\Delta\phi_2) \stackrel{?}{=} E_k(\Delta\phi_2)$, which would say that the system cannot distinguish a positive shape phase from its negative. This would follow if the EOM were invariant under time reversal $t \to -t$: a sign flip of the time variable converts $\cos(2\omega t + \Delta\phi_2)$ into $\cos(-2\omega t + \Delta\phi_2) = \cos(2\omega t - \Delta\phi_2)$, i.e. exactly the sign flip of the phase that we are testing. The conservative ring — the master EOM with $\gamma = 0$ — *is* time-reversal symmetric: $\ddot{x}$, $\omega_0^2 x$, $Lx$, and $\alpha x^3$ are all invariant under $t \to -t$, and a drive that is itself a sum of cosines is unchanged by the reflection. On the conservative ring, Symmetry I would hold exactly.

The damping term $\gamma\dot{x}$ is the only term in Equation 4 that breaks this invariance. A sign flip of time turns $\dot{x}$ into $-\dot{x}$, so the dissipative force changes sign under time reversal — it tells the difference between the system relaxing forward in time and running backward. The dissipative ring is therefore not time-reversal symmetric, which means that Symmetry I does not hold exactly:

$$E_k(-\Delta\phi_2) \neq E_k(\Delta\phi_2) \quad \text{for } \gamma > 0. \tag{19}$$

Sym I is broken in a controlled way — the size of the breaking scales with $\gamma$ and with the cubic-mixing strength that couples the dissipation into the readout. The conservative limit recovers an exact symmetry only in the limit $\gamma \to 0$.

*Why $\phi_0$ is informative.* The two symmetries together determine where in $[0, \pi)$ the peak of $E_5(\Delta\phi_2)$ can sit and what its position carries. Sym II fixes the natural domain to $[0, \pi)$, identifying $\phi_0$ and $\phi_0 + \pi$ as the same observable. If Sym I were also exact, $E_5(\Delta\phi_2)$ would be even in $\Delta\phi_2$, and on $[0, \pi)$ this evenness combined with $\pi$-periodicity would pin the peak of $E_5$ to one of three fixed points: $\Delta\phi_2 = 0$, $\pi/2$, or $\pi$. The observable would then be either degenerate (constant in any parameter we vary externally) or take on at most three discrete values. Sym I being broken is what releases $\phi_0$ from these fixed points and lets it move continuously across $[0, \pi)$ as parameters vary. Figure 5C shows exactly this release in action: $\phi_0$ traces a smooth trajectory from $0.17\pi$ to $0.71\pi$ as $\alpha$ varies from 0.1 to 3.0, occupying intermediate values that no Sym-I-symmetric observable could occupy. The conservative-attractor value $\phi_0 = \pi/2$ at vanishing nonlinearity sits at the unbroken end of the trajectory; moving away from it under increasing $\alpha$ is the substrate recording, in a single number, the joint effect of dissipation and binding on the input shape.

What the broken-symmetry framing buys us, then, is a structural guarantee that $\phi_0$ is informative. Were both symmetries exact, the peak position would be a fixed constant of the substrate, carrying no information about the parameters we vary; were neither symmetry present, $\phi_0$ would still move with parameters but would do so without the natural quotient structure that makes its trajectory clean to read. The asymmetric status — Sym II exact, Sym I broken — is the configuration in which $\phi_0$ is both well-defined as a single number on a clean domain and capable of varying continuously in response to the physics that interests us. The damping that breaks Sym I is what makes the observable do work; this is a general feature of dissipative systems and not specific to the cycle graph or to cubic nonlinearity, and it suggests that trajectories of $\phi_0$ across other physical parameters — drive frequency, coupling strength, the damping itself — carry information of the same kind, structured by the same logic.

# 4 Results

## 4.1 The linear ring as a feature extractor

The first question is whether the ring substrate, in its linear regime, is doing anything useful at all. The framework tells us that it should: the eigenmodes of the cycle graph provide $N$ distinguishable channels into which a temporal input can be sorted, and the Hilbert-envelope readout extracts each channel's amplitude trajectory without requiring the substrate to do any combination of channels with one another. This is bundling, in the precise sense developed in Subsection 3.2. The empirical question is whether this sorting recovers the input's structure cleanly, and whether the resulting representation is more useful than what one would get by reading the input directly. Figures 2 and 3 answer both questions in turn.

Figure 2 compares the linear reservoir's mode-resolved envelope $|a_n(t)|_{\text{env}}$ with a windowed FFT applied directly to the same input signals. Four canonical drives exercise different aspects of the substrate's response. A pure tone at $\omega_5$ produces, in both representations, a narrow band of sustained energy concentrated at the corresponding frequency or mode index — the simplest possible bundling demonstration, with one channel populated and the rest at rest. A linear chirp from $\omega_1$ to $\omega_{12}$ traces a diagonal streak across both panels, showing that as the drive's instantaneous frequency sweeps through the substrate's natural frequencies, the corresponding modes light up in sequence. Each mode's envelope rises and falls as the chirp passes through its resonance, and reading the streak is reading the temporal order in which eigenmodes were excited. A Gaussian burst at $\omega_8$ shows where the two representations begin to part company: the burst is a transient event localised in time, and the reservoir tracks it as a sharp mode-localised pulse with rise and fall times set by the damping, while the windowed FFT smears it into a smudge whose temporal extent is set by the analysis window rather than by the signal's own dynamics. A frequency-modulated tone around $\omega_8$ produces a band of mode amplitude that wobbles in step with the modulation, again visible at higher temporal resolution in the reservoir than in the FFT. Read in the framework's language: bundling distributes the input across distinguishable channels, and each channel's natural timescale $\tau \sim 1/\gamma$ governs how quickly that channel can respond to changes in the input. The reservoir's temporal resolution is the channel's resolution, not an analysis-window choice.

Figure 3 converts the qualitative comparison of Figure 2 into a quantitative one. A four-class problem — noise alone versus a pure tone at one of three well-separated wavenumbers $n \in \{3, 7, 11\}$ — is posed at a sequence of input signal-to-noise ratios spanning $-24$ to 0 dB. For each SNR, the same ridge classifier is trained on two feature vectors of identical dimension: the time-averaged Hilbert envelope of the reservoir's mode amplitudes, and the time-averaged windowed-FFT spectrum of the raw input sampled at the eigenmode frequencies. Both descriptors are sixteen-dimensional and span the same spatial-frequency information; the question is whether passing the signal through the reservoir changes how usefully that information is organised. The answer (Figure 3) is that it does, by a small but consistent margin: the reservoir's psychometric curve is displaced toward the lower-SNR end of the axis by approximately three decibels relative to the FFT baseline, with the displacement visible across the entire transition region from chance to ceiling performance. Both methods recover the signal at high SNR and both fail at sufficiently low SNR; the substrate's contribution is the regime in between, where the reservoir's mode-resolved temporal integration extracts information that the FFT's window-averaged spectrum gives up earlier. This is the bundling architecture's first quantitative payoff: when the channels' integration times are matched to the structure of the signal one is trying to recover, the bundle is more informative than the raw substrate.

What this subsection establishes is that the linear ring is a working feature extractor in the sense of the framework's bundling primitive — the substrate sorts inputs into channels cleanly, and the resulting representation is at least as good as a direct windowed-FFT analysis and modestly better in the regimes where the FFT's fixed-window assumption hurts it. The interesting computational behaviour, however, is what the substrate does once its nonlinearity is turned on: a substrate that only bun-dles sorts information without combining it. The next subsection asks the substrate to do that further job.

### 4.2 Duffing shape sensitivity and the broken-symmetry observable

The Duffing regime turns on the cubic term that the framework identifies as the substrate's binding operation. The empirical question is whether this operation does, in fact, produce a measurable response that depends on input shape rather than on input power alone. A linear system cannot make this distinction — by the closure property discussed in Subsection 3.3, a linear system maps sines to sines and is therefore insensitive to the relative phase among the components of a multi-tone drive, since two drives with identical magnitude spectra produce identical single-frequency outputs at every drive component. The Duffing ring, by mixing input components into harmonics that were not in the drive, gains access to information that the linear ring cannot see. Figure 4 shows this distinction directly, and Figure 5 traces what the resulting shape-dependence looks like across the full domain.

Figure 4 compares the harmonic-energy response of the linear ring (panel B) and the Duffing ring (panel C) to two two-tone drives that differ only in the relative phase of their second-harmonic component, $\Delta\phi_2 = 0$ versus $\Delta\phi_2 = \pi/2$ (panel A). The two drives have identical magnitude spectra: each contains a fundamental at $f_{\mathrm{drive}}$ and a second harmonic at $2f_{\mathrm{drive}}$, in the same proportions. Their waveforms are visibly different — one is amplitude-asymmetric (peaked above the baseline, dipped below), the other is time-asymmetric (sawtooth-like, with a steeper rise than fall) — but a power-spectrum-only analysis treats them as the same signal. The linear ring's harmonic-energy response (panel B) confirms that expectation: $E_1$ and $E_2$ are populated by direct response to the drive's two components, and they are identical to within numerical precision for the two phases. $E_3$ and higher harmonics are at the noise floor of the simulation, because a linear system has no mechanism for generating them. The Duffing ring (panel C) tells a different story. $E_3, E_4, E_5, E_6$ are now visibly populated — the cubic nonlinearity has generated harmonic content that was not in the drive — and the populations differ between the two phases. At $k = 5$, the ratio reaches $E_5(\Delta\phi_2 = \pi/2) / E_5(\Delta\phi_2 = 0) \approx 3.74$. Read in the framework's language: the linear ring bundles, and bundling preserves the input's frequency content channel by channel without combining components. The Duffing ring binds, and binding generates new channels populated by composition products whose existence and magnitudes depend on the relations among the drive's components. The contrast between Figure 4B and C is the empirical signature of binding having occurred.

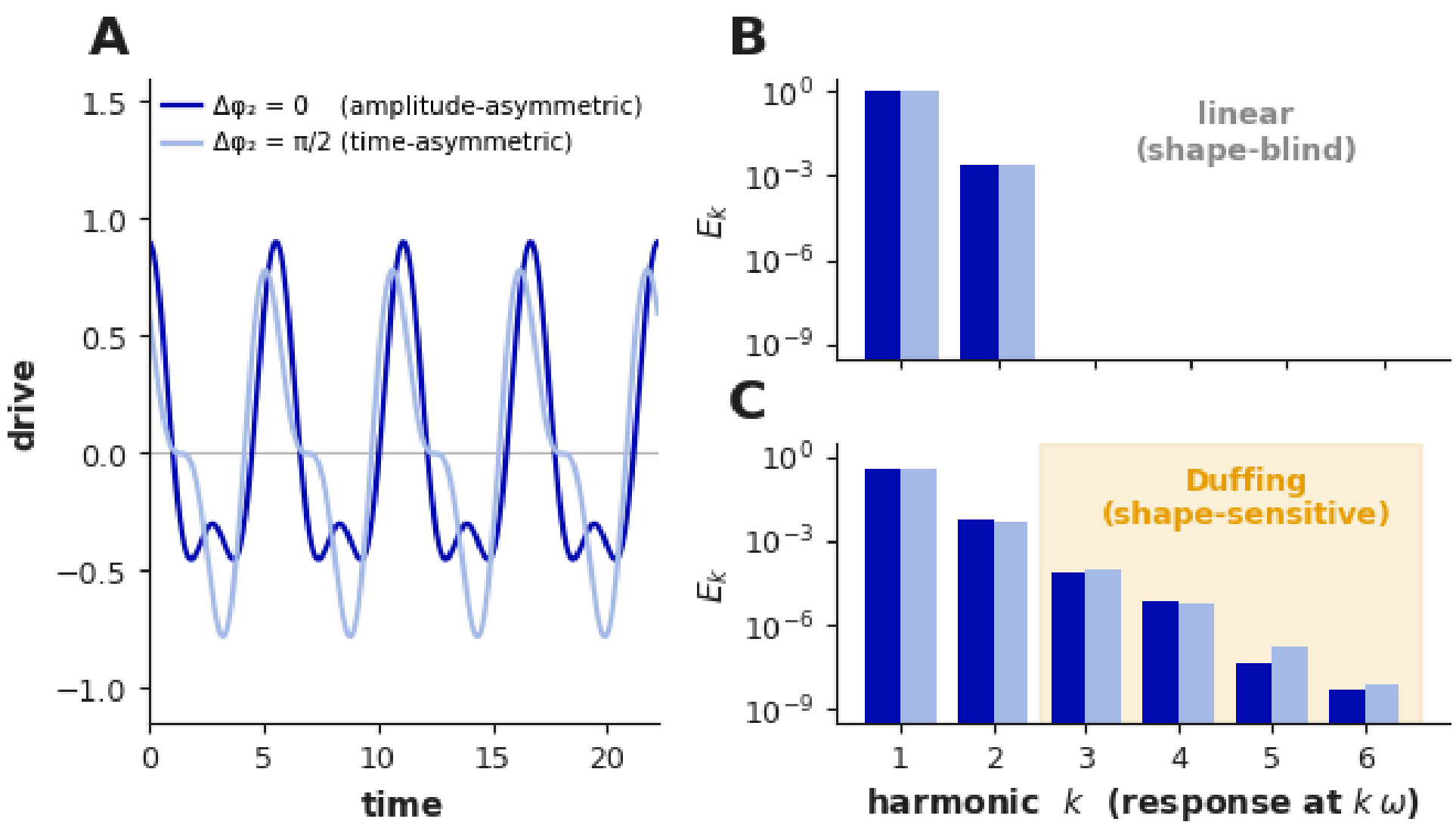

**Figure 4.** Shape sensitivity: the Duffing ring discriminates input waveform shape; the linear ring does not (Duffing-regime parameters: $N = 64, \gamma = 0.15, \omega_0^2 = 1.0, K_c = 0.35, \alpha = 1.5$). (A) Two two-tone drives differing only in the relative phase of the second harmonic: $\Delta\phi_2 = 0$ (amplitude-asymmetric, dark blue) and $\Delta\phi_2 = \pi/2$ (time-asymmetric, light blue). The two drives have identical magnitude spectra $|c_k|^2$. (B) Linear-ring harmonic energies $E_k$ are identical for the two drives — the linear ring is shape-blind. (C) Duffing-ring harmonic energies differ between the two drives, with the largest separation at $k = 5$ ($E_5^B/E_5^A = 3.74$).

Figure 5 extends this two-point comparison to a continuous sweep of the shape phase across the full $\Delta\phi_2 \in [0, 2\pi)$ domain, exposing both waveform symmetries that the framework predicts. Panel A shows $E_5(\Delta\phi_2)$ at the working-point nonlinearity $\alpha = 1.5$, sampled at 64 phase points and interpolated by the trigonometric polynomial of Methods §2.6. The curve has two prominent peaks separated by exactly $\pi$, with identical heights and shapes — the visible signature of Symmetry II. Sym II predicts that $E_5(\Delta\phi_2 + \pi) = E_5(\Delta\phi_2)$ exactly, and the numerical agreement confirms this with relative difference of order $10^{-5}$ between the two halves of the sweep. The two peaks are not two pieces of independent information; they are the same observable read off twice on a domain that double-counts. Identifying them, as the framework's quotient construction $[0, \pi)$ does, turns the two-peak curve into a single peak at $\phi_0 \approx 0.628\pi$.

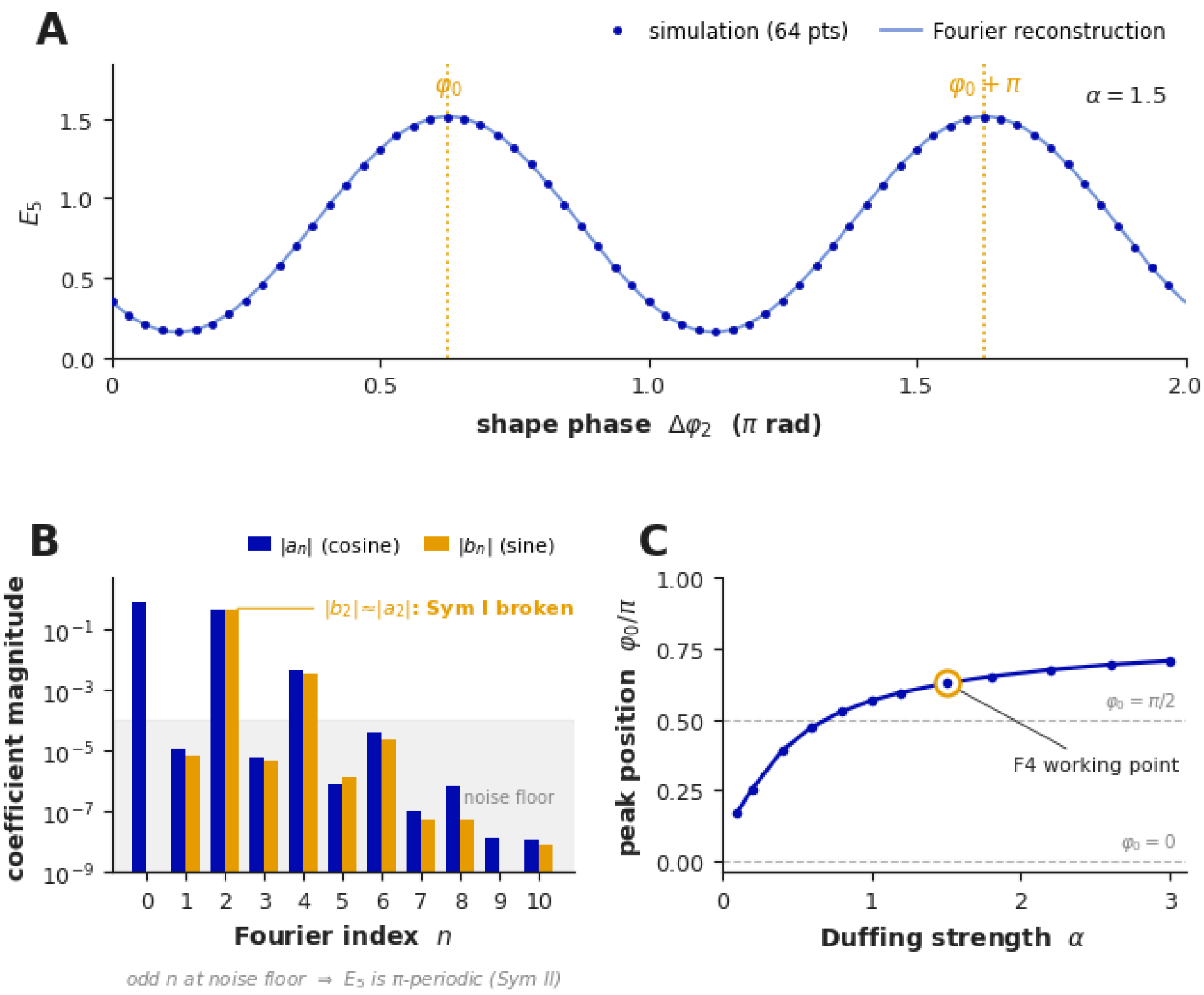

**Figure 5.** Continuous shape-phase sweep at $\alpha = 1.5$ (Duffing regime). (A) Harmonic energy $E_5(\Delta\phi_2)$ over the full sweep $\Delta\phi_2 \in [0, 2\pi)$ (64 points), with Fourier reconstruction (light blue line). The peak at $\phi_0 \approx 0.628\pi$ defines the broken-symmetry shape observable; the second peak at $\phi_0 + \pi$ is identified with the first by Symmetry II. (B) Fourier coefficients $|a_n|$ (cosine, blue) and $|b_n|$ (sine, orange) of $E_5(\Delta\phi_2)$. Odd-$n$ coefficients sit at the noise floor (Symmetry II: $\pi$ -periodicity exact); $|b_2|/|a_2| = 1.000$ (Symmetry I broken at the strongest possible level). (C) Peak position $\phi_0(\alpha)$: $\phi_0$ traverses $0.17\pi \to 0.71\pi$ as $\alpha$ varies $0.1 \to 3.0$. The F4 working point ($\alpha = 1.5, \phi_0 = 0.628\pi$) is marked.

Panel B turns this geometric observation into a numerical one by decomposing $E_5(\Delta\phi_2)$ into its Fourier coefficients $|a_n|$ (cosines, blue) and $|b_n|$ (sines, orange). The pattern across $n = 0$ through 10 is striking and exactly what the symmetry analysis predicts. First, the odd-$n$ coefficients — those that would carry $\pi$-period-violating components — sit at the noise floor of the calculation, several orders of magnitude below the even-$n$ coefficients. This is Sym II, expressed in Fourier language: a $\pi$-periodic function has only even-$n$ Fourier components, and the odd-$n$ floor shows that no $\pi$-period-violating component is present in the data above numerical noise. Second, and more interestingly, the even-$n$ coefficients are split roughly equally between cosine and sine families. At $n = 2$ the ratio is $|b_2|/|a_2| = 1.000$ to three decimal places, the strongest possible breaking of an even-versus-odd distinction in a Fourier series. This is Sym I, broken: a function symmetric in $\Delta\phi_2$ about zero would have only cosine components ($|b_n| \to 0$ for all $n$), and the equality of cosine and sine magnitudes at the dominant mode is the maximally informative deviation from that. Together, the two observations confirm the framework's prediction: at $\alpha = 1.5$, Sym II is exact and Sym I is broken at the largest extent that any observable in this family can show. The damping that breaks Sym I is not subtle in its effect — $\gamma = 0.15$ corresponds to a quality factor of order $\omega_n/\gamma \sim 7$ for the modes near the working drive frequency, hardly an extreme damping — but the nonlinear coupling amplifies its asymmetric signature into the maximum the Fourier basis can register.

Panel C closes the loop by tracing the peak position $\phi_0$ as a function of the nonlinearity parameter $\alpha$. At $\alpha \to 0$, the substrate approaches its conservative-attractor limit and Sym I would become exact, pinning the peak to one of the fixed points $\{0, \pi/2, \pi\}$ that the symmetry argument identified; the trajectory is consistent with this, with $\phi_0$ approaching $\pi/2$ from below as $\alpha \to 0$. As $\alpha$ increases, $\phi_0$ moves continuously across the quotient domain, sweeping from $0.17\pi$ at $\alpha = 0.1$ to $0.71\pi$ at $\alpha = 3.0$ — a range that covers more than half the available domain and is monotone in $\alpha$ over the working range. Read in the framework's language: each value of $\alpha$ defines a different binding operation on the same input, and $\phi_0(\alpha)$ is the substrate recording, in a single number, how the binding algebra responds to changes in the underlying physics. Sym II being exact is what makes $\phi_0$ a clean observable on a single-period domain; Sym I being broken is what releases it from the fixed points and lets it move. The trajectory in panel C is the empirical realization of the "release" framing of Subsection 3.4, and the F4 working point at $\alpha = 1.5$ sits in the middle of the trajectory rather than at any extremum — a moderate nonlinearity that produces a moderately broken symmetry, with $\phi_0$ at $0.628\pi$.

The combined evidence of Figures 4 and 5 establishes the central empirical result of this paper. The Duffing ring binds in a substrate-determined way that the linear ring cannot, and the resulting representation of input shape is structured by two waveform symmetries whose asymmetric status — one exact, one broken — is exactly what makes the broken-symmetry observable $\phi_0$ a meaningful single-number summary of the substrate's response to shape. The remaining empirical question is whether $\phi_0$, defined on clean noise-free data, retains its information content under realistic noise conditions. The next subsection addresses this.

### 4.3 Noise robustness of $\phi_0$

The broken-symmetry observable $\phi_0$ is a derived quantity: a peak position of a Fourier-reconstructed function, computed from harmonic energies, computed from the steady-state response of a nonlinear oscillator network, computed under additive noise. Each step in this chain is a potential point where noise could corrupt the signal and pull $\phi_0$ back toward the conservative-attractor value $\pi/2$ that the framework identifies as the unbroken-symmetry limit. Whether the chain holds together under realistic noise — whether the dissipation-induced shift that we observed cleanly at high fidelity in Subsection 4.2 survives the addition of band-limited Gaussian noise to the drive — is the question that Figure 6 answers.

The experimental setup is straightforward. The two-tone drive is augmented with band-limited Gaussian noise as specified in Methods §2.3, with input signal-to-noise ratios of 30, 20, 10, and 0 dB. At each SNR, eight independent random noise realizations are run, each one driving a fresh sweep of 32 shape-phase points across $[0, 2\pi)$; $\phi_0$ is then estimated independently from each realization's $E_5(\Delta\phi_2)$ curve using the Fourier-reconstruction peak-finder of Methods §2.6. The Duffing parameters are held fixed at the F4/F5 working point ($\alpha = 1.5$, where the noise-free reference value is $\phi_0 = 0.628\pi$).

Figure 6 reports the seed-to-seed mean and standard deviation at each SNR, with individual seeds shown as faint markers behind the aggregate. Two horizontal reference lines mark the relevant landmarks: the noise-free reference at $\phi_0 = 0.628\pi$ (orange dashed) and the conservative-attractor value at $\phi_0 = \pi/2$ (grey dashed), the latter being the value $\phi_0$ would take if Sym I were exact. At 30 dB the seed-to-seed scatter is small and the mean sits at the noise-free reference, as expected for a regime where the noise is well below the signal. As the SNR is lowered to 20 and then 10 dB the scatter grows visibly — some individual seeds excursion to values around $0.4\pi$ and $0.9\pi$ — but the mean stays in the upper half of the quotient domain. Even at 0 dB, where the input noise has the same root-mean-square amplitude as the drive, the mean of $\phi_0$ across eight seeds remains clearly above $\pi/2$ at approximately $0.74\pi$, with the scatter no longer collapsing toward the symmetric limit. What the noise does not do is what one might have most reasonably expected: it does not pull the mean back toward $\pi/2$. Instead, the mean drifts upward, away from $\pi/2$, as the SNR decreases.

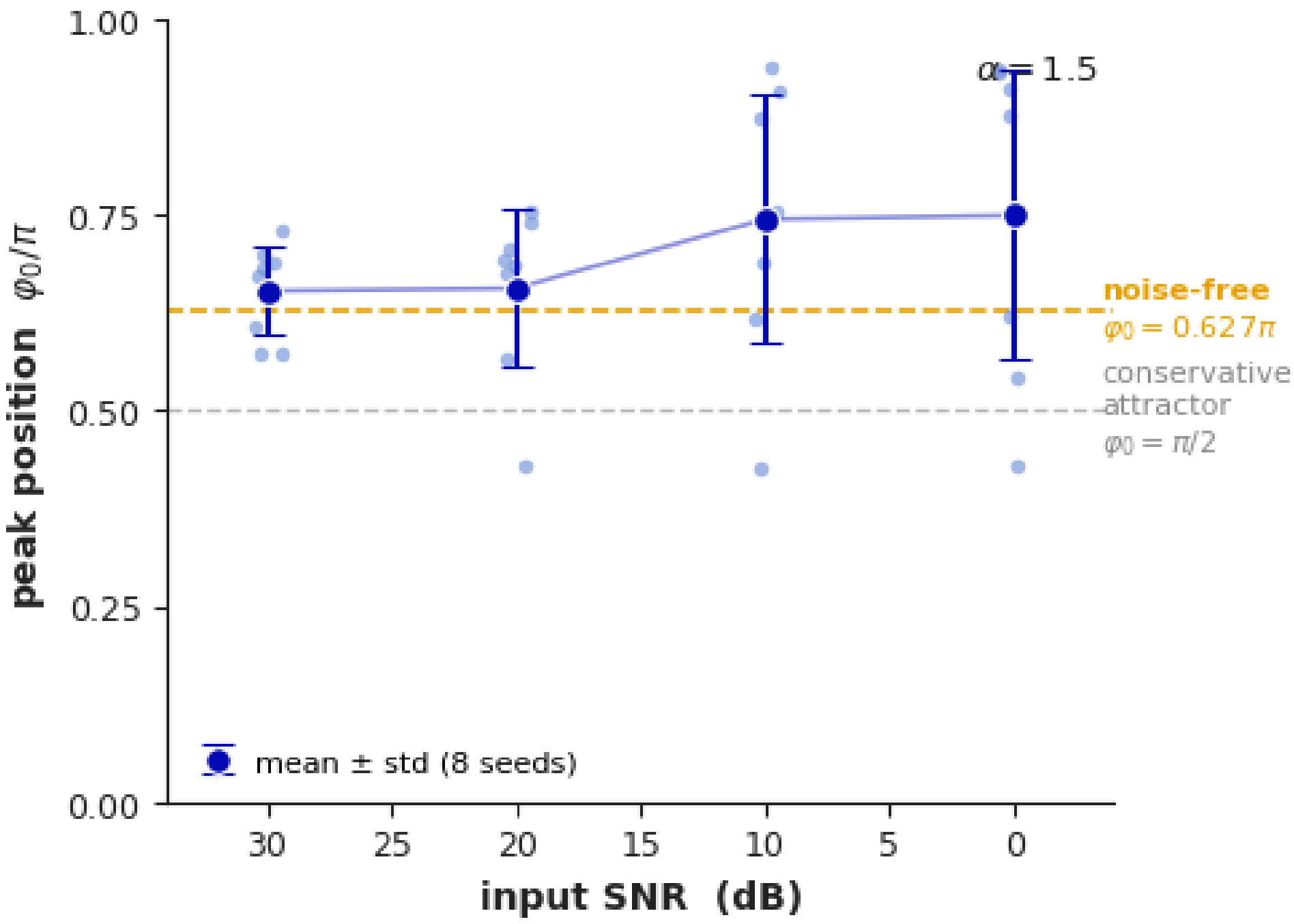

**Figure 6.** Noise robustness of the broken-symmetry observable $\phi_0$ at $\alpha = 1.5$ (Duffing regime). Mean $\pm$ standard deviation across 8 seeds at SNRs of 30, 20, 10, and 0 dB; individual seeds shown as faint dots. Noise-free reference ($\phi_0 = 0.628\pi$, orange dashed) and conservative-attractor limit ($\phi_0 = \pi/2$, grey dashed) are marked. The mean stays clearly above $\pi/2$ down to 0 dB; the single-shot $2\sigma$ detection threshold is approximately 22 dB.

Read in the framework's language, this is the observable's response to having its precision degraded without its symmetry-breaking mechanism being affected. $\phi_0 = \pi/2$ is the value the substrate would produce in the limit of vanishing dissipation — it is the symmetric attractor that Sym I, if exact, would impose. What the simulations do is leave the dissipation $\gamma$ unchanged (at 0.15, the F4 working value) while degrading the input quality. The dissipation that breaks Sym I is therefore still operating on each individual trial, with full strength; what changes is that the noise injects additional energy at frequencies and phases unrelated to the drive's structure, which the cubic mixing then redistributes across the harmonic spectrum. The redistribution does not, however, restore the time-reversal symmetry that the dissipation broke. If anything, it pushes $\phi_0$ further from $\pi/2$, because the noise increases the effective nonlinear driving and the $\phi_0(\alpha)$ trajectory of Figure 5C is monotone increasing in $\alpha$ across the working range.

Quantifying the operational threshold for single-shot detection, we can ask at what SNR a single trial's $\phi_0$ is reliably distinguishable from the symmetric value $\pi/2$. A two-standard-deviation criterion — the mean minus two seed-wise standard deviations stays above $\pi/2$ — is met at approximately 22 dB for these parameter values. Below this threshold, individual trials begin to dip into the range below $\pi/2$ and a single observation cannot reliably distinguish the broken-symmetry signal from a hypothetical symmetric one. The seed-averaged mean, as Figure 6 shows, remains informative considerably further down the SNR axis; the difference is between identifying the broken-symmetry shift on a single noisy trial versus on an ensemble of trials, and the former is the more demanding test.

The empirical robustness shown in Figure 6 closes the loop on the paper's central claim. The Duffing ring binds, the binding produces a representation of input shape that is structured by two waveform symmetries with asymmetric status, and the broken-symmetry observable $\phi_0$ that the asymmetry licenses survives realistic noise without collapsing back toward its symmetric limit. The substrate's bundling-and-binding architecture is therefore not a fragile mathematical curiosity dependent on noise-free conditions; it produces an observable whose informational content is preserved across a substantial range of realistic experimental conditions, and whose failure mode under extreme noise is graceful — a gradual broadening of the seed-to-seed distribution rather than a collapse to the symmetric attractor.

# 5 Discussion

## 5.1 What the paper established

The body has shown three things on a single substrate. First, the cycle graph in its linear regime sorts temporal inputs across distinguishable channels with mode-localised temporal resolution that matches each channel's own dynamics, slightly outperforming a windowed-FFT baseline at low signal-to-noise ratios where the FFT's fixed-window assumption hurts it. Second, switching the substrate into its Duffing regime turns on a mode-mixing operation that the linear ring cannot perform: the same ring now produces harmonic content that depends on the relative phase of the drive's components, not only on their power, and the resulting representation discriminates two-tone drives with identical magnitude spectra by a factor of nearly four in $E_5$. Third, this representation has a clean symmetry structure — a $\pi$-periodicity in the shape parameter that is exact, and a time-reversal symmetry that is broken by the substrate's dissipation — whose asymmetric status licenses a single-number observable, $\phi_0$, that traces a continuous trajectory across the quotient domain as the nonlinearity parameter varies and that remains informative under additive noise down to 0 dB input SNR. The bundling, binding, and broken-symmetry analysis lens of §3 is, in this sense, empirically realised in three distinct figures of the paper.

## 5.2 Bundling and binding in the broader literature

The bundling/binding distinction is not new. It comes from the literature on distributed representations (Schindler and Rahimi 2021; Plate 2003), where the two operations have been the foundational primitives for several decades' worth of work on hyperdimensional computing, vector symbolic architectures, and biologically-inspired representational schemes. What this paper contributes is to apply the distinction as an analytical lens in a setting where it has not, to our knowledge, been previously developed in this form: linear and nonlinear waves on a closed substrate, treated as a single physical system whose parameter regime selects which primitive is being implemented. Three connections to neighbouring literatures are worth making explicit, because the contribution lives at their intersection.

The first connection is to hyperdimensional and vector symbolic computing. In that literature, bundling is typically implemented as additive superposition of high-dimensional vectors, and binding as a chosen algebraic operation — circular convolution in Plate's holographic reduced representations (Plate 2003), component-wise multiplication in other variants — whose specific form is a design parameter of the architecture. On the cycle graph, both operations are inherited from the physics: linear superposition of eigenmodes is the bundling, and the cubic nonlinearity at each node is the binding, with the algebra of admissible bindings (the mod-$N$ selection rule of Equation 9) determined entirely by the substrate's $U(1)$ symmetry and the choice of cubic nonlinearity. The substrate's geometry and dynamics jointly select the binding algebra rather than letting the architect choose it. This shift — from binding-as-design-choice to binding-as-physical-consequence — is what the wave-computing framing offers HD and VSA computing as a complementary perspective.

The second connection is to reservoir computing. The standard framing of reservoir computing posits an input map into a high-dimensional state followed by a linear readout, with the reservoir's job being to supply the dimensionality and the useful nonlinear features that downstream linear readouts can pick up (Jaeger 2001; Maass, Natschläger, and Markram 2002; Lukoševičius and Jaeger 2009; Enel et al. 2016; Maksymov 2024; Gallou et al. 2024). Reading the same architecture through the bundling/binding lens gives a finer resolution: bundling provides the dimensionality (the substrate's $N$ distinguishable channels), binding provides the compositional structure (the harmonic enrichment that depends on input shape rather than only on input power), and the linear readout is the operation that picks compositional content out of a bundled representation. The two regimes of the ring substrate serve as a clean separation of the two contributions: F2 and F3 quantify what bundling alone delivers, F4 through F6 quantify what binding adds.

The third connection is to the modelling of oscillatory neural dynamics with continuous neural fields and discrete eigenmode decompositions (Nunez 1974; Robinson, Rennie, and Wright 1997; Müller et al. 2022; Xie et al. 2021; Pang et al. 2023; Glomb et al. 2020). Mode decomposition has been used in cortical modelling for several decades — as a basis for analytical tractability, as a description of large-scale spatial coherence patterns, and more recently as a concrete framework for relating EEG/MEG observables to underlying field dynamics. What the wave-computing framing adds to this tradition is a computational reading: the eigenmodes are not merely a mathematically convenient basis but a substrate's bundling architecture, and the nonlinear couplings among modes are not merely a complication of the dynamics but the substrate's binding operations. Whether real cortical dynamics can usefully be read this way is an empirical question that this paper does not address; what the paper does establish is that the formal apparatus linking eigenmode dynamics to bundling/binding computation works on the simplest possible non-trivial substrate and respects the symmetry constraints one would expect.

A fourth connection is to the engineering tradition of computing with networks of coupled oscillators — oscillatory neural networks, oscillator-based Ising machines, and related architectures whose lineage runs from the parametron through Hoppensteadt and Izhikevich's oscillatory neurocomputer to current CMOS, spintronic, and phase-transition-device implementations (Hoppensteadt and Izhikevich 1999; Todri-Sanial et al. 2024). The dominant formalism there is Kuramoto's phase reduction, in which the oscillators are pointlike nodes with weak pairwise coupling, amplitudes are constant, and information is encoded in phase differences between nodes. The cycle graph studied here meets that tradition in a specific limit: at small drive amplitudes the mode amplitudes evolve slowly compared to their oscillation periods, and a standard reduction recovers a Kuramoto-like description on the substrate's eigenmodes. The Duffing regime sits outside that limit. Amplitudes vary, the inter-mode coupling is generated by the substrate's own cubic nonlinearity rather than supplied externally as a designed coupling matrix, and the bound representation lives in the harmonic spectrum produced at each location rather than in pairwise phase relations between fixed nodes. What the wave-computing framing offers oscillator-network computing as a complement is a continuum, shape-based regime in which the substrate's geometry, rather than a designer's choice of coupling, fixes the algebra of admissible bindings.

### 5.3 What generalises and what is substrate-specific

It is worth being explicit about which features of the analysis are properties of the cycle graph specifically, and which are expected to carry over to any closed wave substrate carrying similar algebraic structure. Three threads are worth distinguishing.

Bundling is general. Any closed wave substrate with a discrete eigenspectrum sorts temporal inputs onto its eigenmodes by linear superposition, with the eigenmodes' natural frequencies determining which input components couple most strongly to which modes. The cycle graph's particular Fourier eigenmodes are specific to its topology, but the operation of populating distinguishable channels and reading them out via a linear projection works on any substrate whose linearised EOM diagonalises in some basis. The Hilbert-envelope readout we used, the harmonic-energy summary $E_k$, and the windowed-FFT baseline comparison are all substrate-independent in this respect. What changes from substrate to substrate is the structure of the eigenmodes themselves: the cycle graph offers two-fold cosine/sine degeneracy at every non-trivial wavenumber and a single $U(1)$ phase angle per mode pair; substrates carrying richer continuous symmetries offer correspondingly richer multiplet structure and additional parameters per multiplet.

Binding is substrate-specific. The cubic term $\alpha x_i^3$ generates harmonic content on any substrate, but the algebra of admissible cubic mode couplings — which input modes can mix into which output modes — is determined by the substrate's symmetries. On the cycle graph this algebra is the mod-$N$ selection rule of Equation 9: a sparse, integer-arithmetic constraint on triples of wavenumbers, easily computed and easily interpreted. On substrates with richer continuous symmetries the analogous constraint takes a more elaborate form, with non-trivial coupling coefficients (Clebsch-Gordan-type structure) replacing the Boolean mod-$N$ test, and with the resulting bound representations carrying compositional content of correspondingly higher descriptive power. The framework developed here scales: it should describe binding on any closed substrate, but the algebra to be solved is what each substrate gives.

The broken-symmetry observable is structurally general but locally proven. The asymmetric-symmetries argument that licenses $\phi_0$ uses two facts about the master EOM: that the EOM and the readout are both even in $x_i$ (Sym II), and that the dissipation term is the sole obstruction to time-reversal invariance (Sym I). Neither fact is specific to the cycle graph; both should hold for any oscillator network whose master EOM has the same algebraic structure. What we proved on the ring is therefore expected to extend to other closed substrates with the same EOM structure, with the exact periodicity of $E_k$ in the shape parameter, the exact symmetric-attractor value of $\phi_0$, and the parameter trajectory across the quotient domain all reading off the same algebra. The proofs are local to the ring; the argument that licenses them is not.

### 5.4 Limitations and open questions

The paper analyses one closed substrate, in two parameter regimes, on synthetic signals constructed to exercise specific features of its response. This is, by design, a narrow empirical scope. Three categories of open question follow.

The first concerns substrates. The bundling/binding/symmetry analysis developed here uses the cycle graph as a vehicle, but the analysis itself is intended to generalise to richer closed substrates. What features of the analysis carry over, what features need substrate-specific modification, and what new phenomena appear only on substrates with richer symmetries are open questions that the paper provides apparatus for asking but does not itself ask.

The second concerns drives. The two-tone shape probe used in Figures 4–6 exercises the substrate's response to a one-parameter family of input waveforms with a single fundamental frequency and one shape-controlling phase. Multi-frequency drives, time-varying drives, and waveforms with internal modulation structure are not addressed; each would require a redoing of the symmetry analysis with the specific structure of the new drive in mind. The framework should accommodate them, but each extension is its own analytical project.

The third concerns the connection to real signals. This paper does not analyse any biological time series. The bundling, binding, and broken-symmetry framework was developed with biological oscillatory dynamics — including electroencephalographic signals, in which non-sinusoidal waveform structure is known to carry physiologically meaningful information (Cole and Voytek 2017) — as part of the broader motivation, but the empirical step of asking whether $\phi_0$ or related observables capture useful information about real cortical waveform shape is left for elsewhere. The empirical neuroscience analogues of the cubic mode-mixing analysed here — cross-frequency coupling between distinct populations and the higher-order spectral signatures (bispectrum, bicoherence) of waves whose harmonics carry informative phase relations (Canolty and Knight 2010; Hyafil 2015) — are natural connection points for that future work. A parallel and more direct measurement tradition addresses the same non-sinusoidal structure in the time domain rather than the frequency domain, by parsing oscillatory signals into individual cycles and measuring per-cycle amplitude, period, and rise–decay or peak–trough asymmetry (Cole and Voytek 2019; Cole 2018; Cole et al. 2017). The time-domain cycle-feature description and the substrate-level harmonic-energy description developed here address overlapping aspects of the same non-sinusoidal structure from complementary directions — one as a measurement made directly on the signal, the other as a physical model of how such structure can arise on a substrate that bundles linearly and binds nonlinearly — and articulating the formal correspondence between them is a natural separate exercise. The empirical motivations for this framework also come from the author's own prior work on iEEG signal analysis. Eigenvalue-spectrum analysis of the multichannel iEEG correlation matrix has shown stereotyped reorganisations of the spectrum across focal-onset seizures, with secondary generalisation accompanied by transient decorrelation followed by re-concentration of variance onto a few dominant modes (Schindler et al. 2007; Schindler, Elger, and Lehnertz 2007); the methodological pairing of this multivariate reading with single-signal symbolic-dynamics characterisation of each channel (Schindler et al. 2012) drew an explicit distinction between features extracted locally and compositional structure read across channels — the empirical precursor to the bundling/binding lens the present paper develops formally. Those same stationary correlation patterns, complementarily, admit a reading as standing waves of cortical activity (Müller et al. 2014), which a substrate of the present kind, on a topology with appropriate spatial structure, would be expected to support directly. Saw-tooth time-irreversible iEEG signals — the asymmetric rise/decay structure that the broken time-reversal symmetry of $\phi_0$ formalises in the substrate's response — have subsequently been shown to mark the epileptogenic zone (Schindler et al. 2016), and the seizure itself has been analysed as the collapse of a chimera-like coexistence of synchronised and desynchronised cortical populations (Andrzejak et al. 2016) — a symmetry-restoration phenomenon whose substrate-level analogue within the present framework is the collapse of the eigenmode bundling onto a small number of dominant modes. Ongoing efforts in ultra-long-term ambulatory EEG monitoring (Yilmaz et al. 2024) provide a complementary empirical platform on which scaled-up versions of the present framework can in principle be tested. None of these connections are claimed as results of the present paper; they are stated to make explicit the empirical questions the apparatus is intended to be eventually capable of asking. What this paper provides is the apparatus for asking that question on a substrate where the analysis is fully transparent; the asking, and the substrate-modelling work that would be needed for the answer to bear on cortical interpretation, is a separate body of work.

What the paper does is to develop the bundling and binding distinction into an analysis of waves on a closed substrate and to identify a single observable, structured by a broken symmetry of the substrate's dissipation, that can summarise the bound representation of input shape under realistic noise. What it does not do — and the body has been deliberate in not attempting — is to claim that this observable is biologically grounded, clinically applicable, or definitively superior to specialised shape-detection methods on application data. Whether any of these stronger claims hold is a question for the work that follows.

## 6 Conclusion

A linear ring sorts a temporal input across distinguishable channels; adding a smooth nonlinearity makes the same ring compose those channels into outputs whose informational content depends on the relations among the input's components rather than on their amplitudes alone.

Reading the substrate's response through the bundling, binding, and broken-symmetry analysis lens turns this observation into a quantitative architecture: a single observable, $\phi_0$, licensed by the asymmetric status of two waveform symmetries — one exact, one broken by the substrate's dissipation — summarises the bound representation of input shape on a clean quotient domain and remains informative under realistic noise. The cycle graph, the simplest closed substrate carrying a continuous symmetry, is sufficient to develop and demonstrate the framework; extending it to richer substrates, more elaborate drives, and real biological signals are open questions for the work that follows.

## Data and Code Availability

All code required to reproduce Figures 1–6 of this paper is publicly available at https://github.com/KasparSchindler449/wavecomputing-paper1 under the MIT licence. The exact version of the code used to produce the figures in this manuscript is archived at Zenodo, DOI 10.5281/zenodo.20055786, released as tag `v1.0.0` of the repository.

No external data were analysed in this study; all signals are synthetically generated by the code released alongside the manuscript.

The figures in this manuscript were produced in Google Colab Pro using NumPy 2.0.2, SciPy 1.16.3, and Matplotlib 3.10.0. These versions are pinned in `requirements.txt` of the released code archive. Random seeds for all stochastic figures are hardcoded in the source so that re-runs from a fresh clone reproduce the reported figures bit-identically.

## Acknowledgments

The author thanks the open-science communities that made this work possible, in particular the developers of the open scientific Python stack (NumPy, SciPy, matplotlib, JAX) on which all simulations and figures were built.

Claude (Anthropic, principally Claude Opus 4.7) was used to support writing, coding, consistency checks, and copy-editing; all conceptual choices, mathematical content, simulations, figures, and final editorial decisions are the author's.

This work received no external funding.

The author declares no competing interests.

## G The cubic coupling tensor on the ring

This appendix derives the explicit form of the cubic mode-coupling tensor $T_{n\,m_1 m_2 m_3}$ that appears in the eigenmode-space Duffing equation, Equation 8, and the selection rule of Equation 17 in the process. The aim is didactic: every step is shown, and at each step the algebraic content is paired with a physical picture, so that a reader who has not seen this kind of calculation before can follow the argument without filling in implicit moves.

### What we are computing, and why

The master equation of motion (Equation 4) lives in node coordinates $\{x_i(t)\}_{i=0,\ldots,N-1}$. Four of its five terms are linear in $x_i$: the inertia, the on-site stiffness, the damping, and the Laplacian coupling. These four terms diagonalise on transformation to the eigenmode basis (§2.2, Equation 4), giving an EOM in which each mode evolves independently. The fifth term, the cubic nonlinearity $\alpha\, x_i^3$, does not diagonalise: it couples eigenmodes to one another, and the question of *which* eigenmodes it couples is what this appendix answers.

The plan is to take the cubic term as it appears at each node, and ask what its projection onto eigenmode $n$ looks like when the node displacements are themselves a sum over eigenmodes. The result is a sum over triples of input modes, with each triple multiplied by a coupling coefficient. The coefficient is the cubic coupling tensor $T_{n\,m_1 m_2 m_3}$. Most coefficients turn out to be zero; the ones that are not zero are exactly those whose mode indices satisfy the selection rule of Equation 17.

### Setting up the basis change

We work in the complex-valued Fourier basis introduced in Methods §2.1. For each eigenmode $n \in \{0, 1, \ldots, N-1\}$, the basis vector has components

$$[\boldsymbol{v}_n]_j = \frac{1}{\sqrt{N}} e^{2\pi i\, nj/N}, \quad j = 0, \dots, N-1. \tag{20}$$

A node displacement $x_j(t)$ can be written as a superposition of mode amplitudes,

$$x_j(t) = \frac{1}{\sqrt{N}} \sum_{m=0}^{N-1} u_m(t)\, e^{2\pi i\, mj/N}, \tag{21}$$

where the $\{u_m(t)\}$ are the complex-valued mode amplitudes. For the displacement at any single node to be real, the mode amplitudes must satisfy a reality condition: $u_{N-m} = u_m^*$, the complex conjugate, for every $m \neq 0$ (and $u_0$ itself real). This is what allows the $N$ complex amplitudes to encode only $N$ real degrees of freedom rather than $2N$.

## Cubing a single node displacement

We now compute $x_j^3$ by substituting Equation 21 into itself three times:

$$x_j^3 = \frac{1}{N^{3/2}} \sum_{m_1, m_2, m_3 = 0}^{N-1} u_{m_1} u_{m_2} u_{m_3}\, e^{2\pi i\,(m_1+m_2+m_3)\,j/N}. \tag{22}$$

This is the entire content of "cubing a sum of waves": each term in the triple sum is a product of three mode amplitudes, multiplied by an exponential whose phase is the *sum* of the three mode wavenumbers, evaluated at node $j$. The physical content is straightforward. Three waves of wavenumbers $m_1, m_2, m_3$ multiplied together produce, at every point in space (every node $j$), an oscillation whose spatial wavenumber is $m_1 + m_2 + m_3$. This is harmonic mixing in the spatial domain: just as the product of two cosines at different frequencies produces sum and difference frequencies in the temporal domain, the product of three plane waves at different spatial wavenumbers produces a wave at their summed spatial wavenumber.

## Projecting onto a single output mode

To recover the projection of $x_j^3$ onto eigenmode $n$, we apply the inverse transform of Equation 21 to the cubed displacement:

$$\left[\boldsymbol{V}^T\!\left(x_j^3\right)\right]_n = \frac{1}{\sqrt{N}} \sum_{j=0}^{N-1} e^{-2\pi i\, nj/N}\, x_j^3. \tag{23}$$

Substituting Equation 22 and collecting like terms, the $j$-summation factors out and contains the entire selection-rule information:

$$\frac{1}{N} \sum_{j=0}^{N-1} e^{2\pi i\,(m_1+m_2+m_3-n)\,j/N} = \begin{cases} 1 & \text{if } m_1 + m_2 + m_3 \equiv n \,(\text{mod}\, N), \\ 0 & \text{otherwise.} \end{cases} \tag{24}$$

Equation 24 is the pivot of the whole calculation, and it has a clean physical reading. The sum over $j$ is asking whether the spatial wave at wavenumber $m_1 + m_2 + m_3$ that the cube produced has any overlap with the spatial wave at wavenumber $n$ that we are projecting onto. The two waves have full overlap if their wavenumbers agree (modulo $N$, because the ring is periodic and waves at $N$ and 0 are the same wave) and zero overlap otherwise. This is the precise mechanism by which the closed-substrate self-encounter that we discussed in §3.3 produces a coherent, selection-rule-respecting binding rather than a smeared-out broadband response.

### The cubic coupling tensor

Combining Equation 22, Equation 23, and Equation 24, the projection of the cubic node nonlinearity onto eigenmode $n$ becomes

$$\left[\boldsymbol{V}^T\left(x_j^3\right)\right]_n = \frac{1}{\sqrt{N}} \sum_{m_1,m_2,m_3} u_{m_1} u_{m_2} u_{m_3}\, \delta[m_1 + m_2 + m_3 \equiv n\,(\mathrm{mod}\,N)], \quad (25)$$

where $\delta[\cdot]$ is the Kronecker delta on the modular condition (1 if the congruence holds, 0 otherwise). Identifying this with the form of Equation 8, the cubic coupling tensor is

$$T_{n\,m_1 m_2 m_3} = \frac{1}{\sqrt{N}}\, \delta[m_1 + m_2 + m_3 \equiv n\,(\mathrm{mod}\,N)]. \quad (26)$$

The tensor is sparse: most of its $N^4$ entries are zero, and the non-zero entries are all equal in magnitude. What carries the substrate's structure is the pattern of zeroes, not the magnitudes; this is what is meant by "the substrate determines the binding algebra without choosing the coefficients."

### The selection rule with sign freedom

The selection rule as stated in Equation 24 is "$m_1 + m_2 + m_3 \equiv n$" — all signs positive on the left. The selection rule as stated in §2 and §3 (Equation 9 and Equation 17) reads "$\pm m_1 \pm m_2 \pm m_3 \equiv n$" — with all sign combinations allowed. The two are not in conflict: the sign freedom comes from the reality condition $u_{N-m} = u_m^*$ that the node displacements must be real.

A mode amplitude $u_{N-m}$ is the complex conjugate of $u_m$, which in physical terms means it represents a wave of the same shape travelling in the opposite direction. The cube can therefore reach any output mode $n$ via any combination of three mode amplitudes whose wavenumbers sum to $n$ modulo $N$ *when each input wavenumber is allowed to be either $m$ or $N - m$* — equivalently, either $+m$ or $-m$ in the modular arithmetic. Hence the sign-free form: $\pm m_1 \pm m_2 \pm m_3 \equiv n\,(\mathrm{mod}\,N)$.

In the real-valued cosine/sine basis used elsewhere in the paper, this manifests as eight independent sign combinations per triple of unsigned wavenumbers, each connecting a specific cosine/sine parity pattern in the inputs to a specific cosine/sine parity in the output. The detailed bookkeeping is mechanical and adds nothing to the substantive content; what matters is that the modular arithmetic and the sign freedom together give the substrate's full algebra of admissible cubic bindings.

### Reading the selection rule

A useful way to internalise Equation 24 is by example. Consider a drive that excites only modes $m = 1$ and $m = 2$. Cubing the resulting node displacement and projecting onto each output mode, we ask which modes can possibly be populated by the cubic nonlinearity. The triples $(m_1, m_2, m_3)$ available to us are limited to $\{1, -1, 2, -2\}$ (using the sign-free form), and their summed wavenumbers modulo $N$ are $\{0, \pm 1, \pm 2, \pm 3, \pm 4, \pm 5, \pm 6\}$ before considering modular wrap-around. For the working-value $N = 64$, these all sit well inside the modular range, so the predicted populated modes are at wavenumbers $\{0, 1, 2, 3, 4, 5, 6\}$ — which is exactly the set of harmonics we observe in Figure 4C. Higher harmonics ($k = 7, 8, \ldots$) require fifth-order or higher products in the equation of motion to populate; the cubic term cannot reach them.

This worked example also clarifies a feature of the Duffing ring's response that might otherwise seem mysterious: the selection rule predicts harmonic content up to $3 \times \max(m)$ where $\max(m)$ is the largest mode wavenumber in the drive, and this is what F4 panel C shows. The substrate's sparseness is therefore not a limitation but a feature: it tells us in advance which modes the binding can populate and which it cannot, enabling a clean separation between the information the cubic produces and the information that would require richer (higher-order, or different-substrate) operations.